\documentclass[cameraready]{template/mac_automl_xmu_blue}

\graphicspath{{content/}}

\usepackage[toc,page,header]{appendix}

\usepackage{xargs}
\usepackage{todonotes}
\usepackage{tabularx}
\usepackage{xspace}

\newcommand{\ours}{%
  \textcolor[RGB]{30,58,138}{T}%
  \textcolor[RGB]{43,55,151}{i}%
  \textcolor[RGB]{56,52,164}{m}%
  \textcolor[RGB]{70,49,178}{e}%
  \textcolor[RGB]{83,46,191}{P}%
  \textcolor[RGB]{96,43,204}{L}%
  \textcolor[RGB]{109,40,217}{E}\xspace}
\newcommand{\gain}[1]{\textcolor{green!50!black}{\ensuremath{(\uparrow #1)}}}
\newcommand{\dropv}[1]{\textcolor{red!65!black}{\ensuremath{(\downarrow #1)}}}
\newcommand{\same}[1]{\textcolor{gray!50!black}{\ensuremath{(#1)}}}

\newtcolorbox{findingbox}[1][Findings]{
  enhanced,
  sharp corners,
  colback=black!2,
  colframe=black!75,
  boxrule=0.8pt,
  coltitle=black!85,
  fonttitle=\bfseries\itshape,
  attach boxed title to top left={xshift=1em,yshift=-\tcboxedtitleheight/2},
  boxed title style={
    sharp corners,
    colframe=white,
    colback=white,
    boxrule=0pt,
    left=2pt,right=2pt,top=0pt,bottom=0pt
  },
  title=#1,
  left=6pt,right=6pt,top=10pt,bottom=6pt
}

\setleftheadercontent{%
  \headerlogospace{1.6mm}%
  \headerlogo[-0.08\height]{15.2mm}{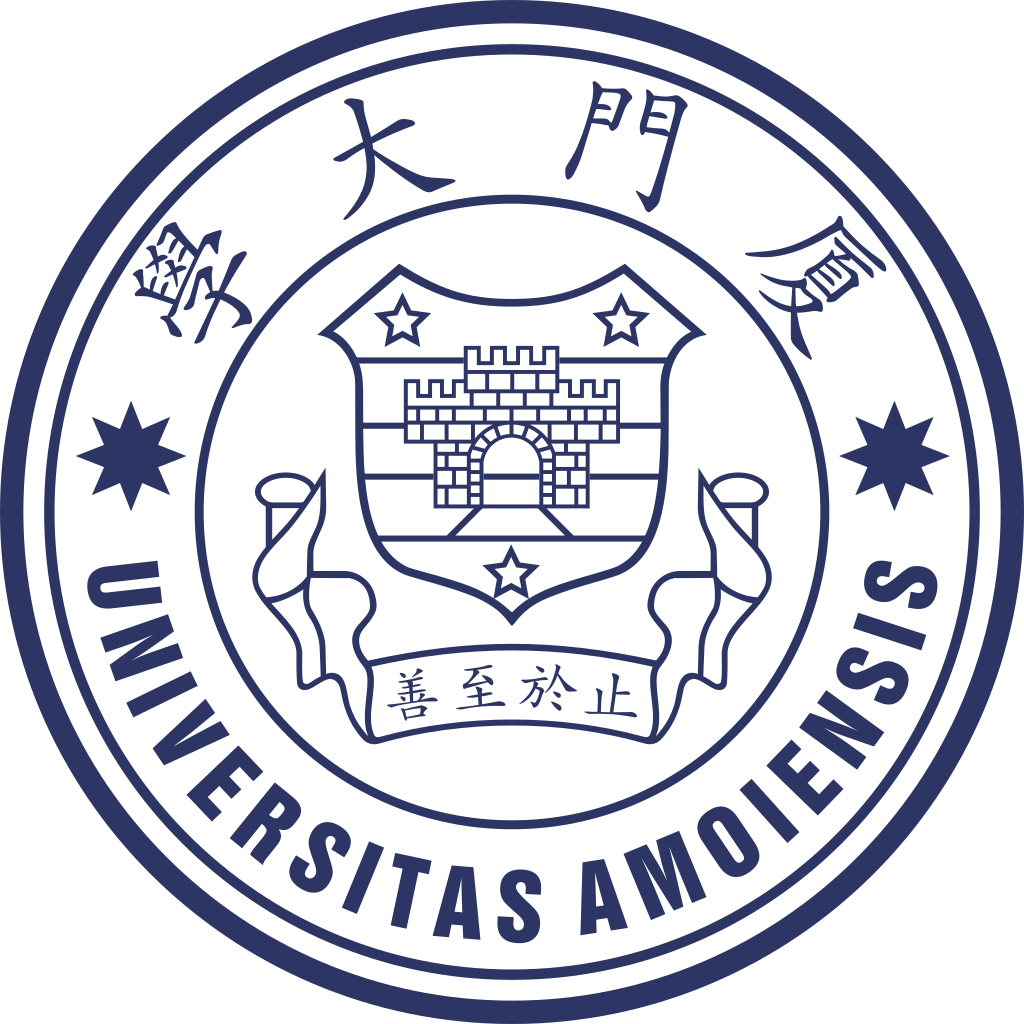}%
  \headerlogospace{2.8mm}%
  \headerlogo[-0.06\height]{12.6mm}{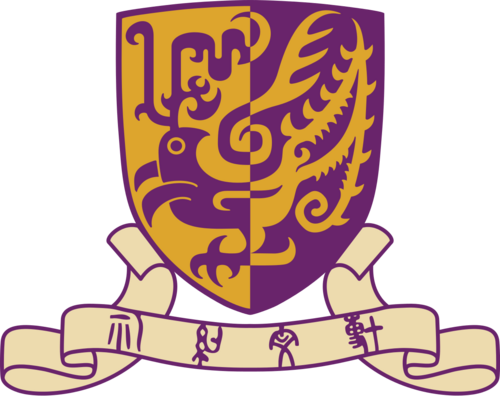}%
  \headerlogospace{2.8mm}%
  \headerlogo[-0.06\height]{12.6mm}{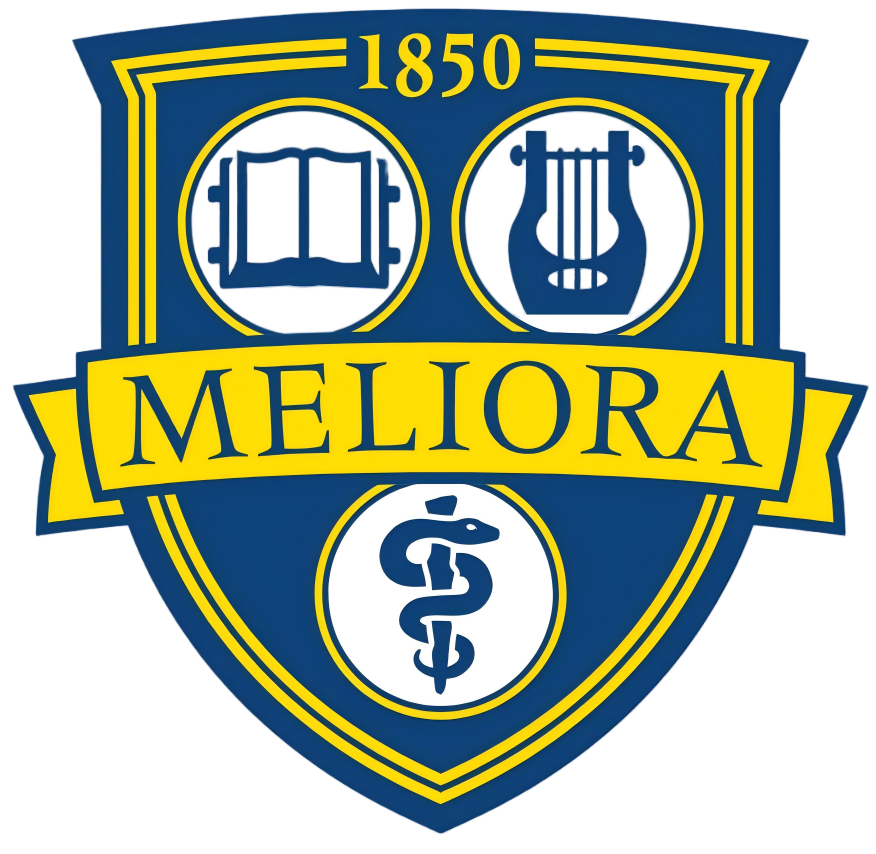}%
  \headerlogospace{2.8mm}%
  \headerlogo[-0.05\height]{12.8mm}{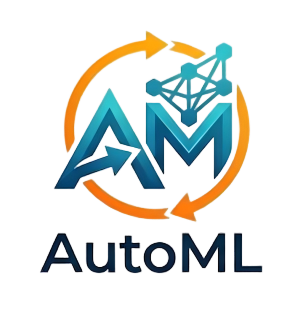}%
  \headerlogospace{2.8mm}%
  \headerlogo[-0.04\height]{8.2mm}{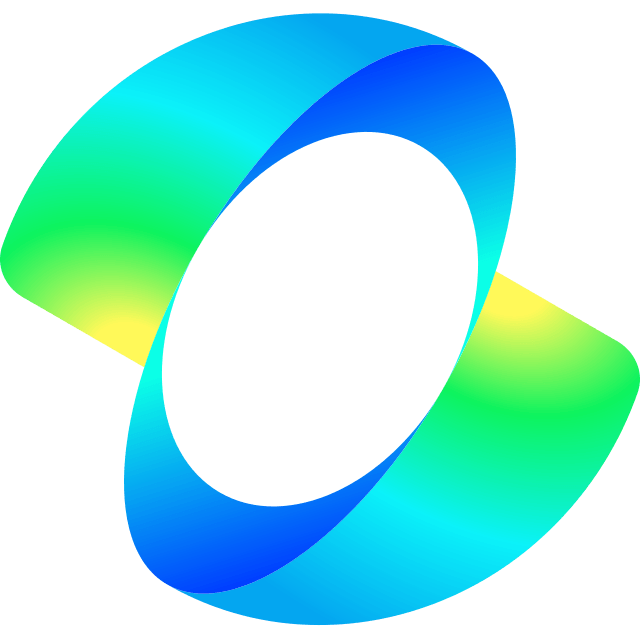}%
  \headerlogospace{2.05mm}%
  \headerlogo[-0.04\height]{6.56mm}{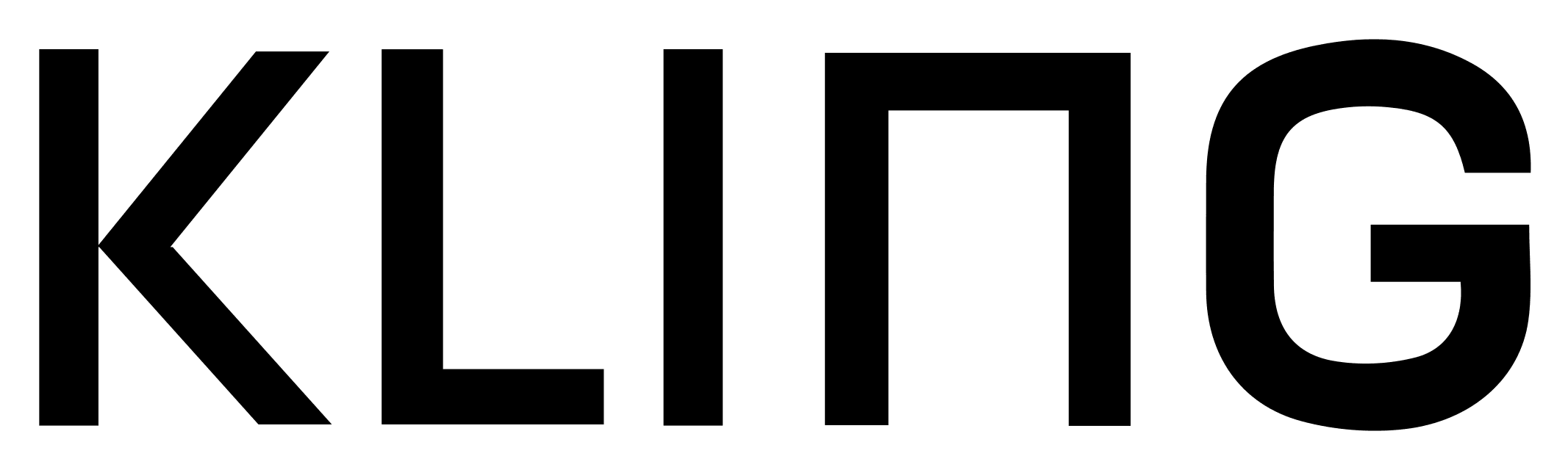}%
}

\setrightheadericon{%
  \headerlogo[-0.06\height]{13.0mm}{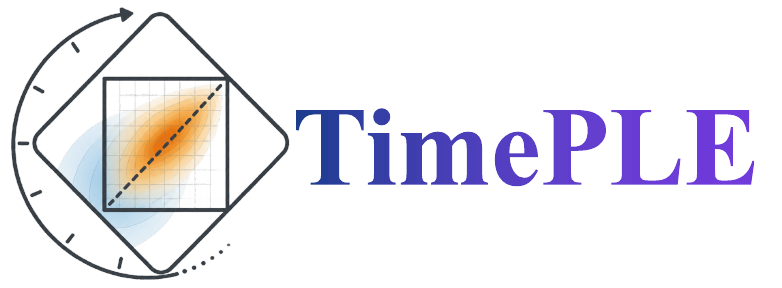}%
}

\setrunningheadericon{%
  \headerlogospace{1.2mm}%
  \headerlogo[-0.03\height]{8.2mm}{assets/branding/automl.png}%
}

\setheadergroupname{MAC-AutoML}

\title{\ours: Rethinking Temporal Representation for Video Temporal Grounding}

\newcommand{\equalcontribmark}{\ensuremath{\ast}}
\newcommand{\internshipmark}{\ensuremath{\dagger}}
\newcommand{\projectleadermark}{\ensuremath{\ddagger}}

\newcommand{\equalcontribcontactmark}{%
  \textsuperscript{\(\ast\)}%
}

\newcommand{\internshipcontactmark}{%
  \textsuperscript{\(\dagger\)}%
}

\newcommand{\projectleadercontactmark}{%
  \textsuperscript{\(\ddagger\)}%
}

\newcommand{\frontaffiliationfont}{%
  \small
  \rmfamily
  \mdseries
  \upshape
}

\newcommand{\frontcontactfont}{%
  \footnotesize
  \rmfamily
  \mdseries
  \upshape
}

\setfrontauthors{%
  \authorrow{%
    \authorentry
      {Yuhui Zeng}
      {1,4,\equalcontribmark\internshipmark},%
    \authorsep
    \authorentry
      {Xinyu Mao}
      {2,4,\equalcontribmark\internshipmark},%
    \authorsep
    \authorentry
      {Xiaokun Liu}
      {4,\projectleadermark\corremailmark},%
    \authorsep
    \authorentry
      {Xin Tao}
      {4}%
  }%
  \authorrow{%
    \authorentry
      {Jinfa Huang}
      {3},%
    \authorsep
    \authorentry
      {Jiayi Ji}
      {1},%
    \authorsep
    \authorentry
      {Xiawu Zheng}
      {1,\corremailmark}%
  }%
}

\setfrontaffiliations{%
  \affiliationline{%
    {\frontaffiliationfont
      \authormark{1}%
      Media Analytics and Computing Lab, Xiamen University,
      Xiamen, China%
    }%
  }%
  \affiliationline{%
    {\frontaffiliationfont
      \authormark{2}%
      The Chinese University of Hong Kong, HKSAR, China%
    }%
  }%
  \affiliationline{%
    {\frontaffiliationfont
      \authormark{3}%
      Department of Computer Science, University of Rochester,
      Rochester, NY, USA%
    }%
  }%
  \affiliationline{%
    {\frontaffiliationfont
      \authormark{4}%
      Kling Team, Kuaishou Technology, China%
    }%
  }%
}

\setfrontcontact{%
  \contactline{%
    {\frontcontactfont
      \equalcontribcontactmark\,
      \textbf{Equal Contribution}%
    }%
  }%
  \contactline{%
    {\frontcontactfont
      \projectleadercontactmark\,
      \textbf{Project Leader}%
    }%
  }%
  \contactline{%
    {\frontcontactfont
      \corremailmark\,
      \textbf{Corresponding Authors}%
    }%
  }%
  \contactline{%
    {\frontcontactfont
      \internshipcontactmark\,
      This work was conducted during the authors' internships
      at Kling Team.%
    }%
  }%
}

\usecustomauthorlayout

\abstract{\begin{abstract}

Video temporal grounding (VTG) aims to localize the continuous video interval described by a natural-language query.
However, current VLM-based methods typically produce this interval indirectly through two endpoint outputs, represented either as discrete timestamp tokens or continuous boundary coordinates. 
These formulations differ in how endpoints are encoded, but not in what is predicted: the event interval remains a derived object, while interval validity, duration, and interval-level similarity are handled only implicitly. 
We propose \textbf{\ours}, which reformulates VTG from endpoint prediction to interval-native grounding by predicting a single joint distribution over valid temporal intervals. 
\ours maps each interval to a point in a canonical position–duration square, where every support point corresponds to a valid span and neighboring points represent geometrically similar intervals. 
Given a video and query, the VLM generates a single latent <|TIMESPAN|> token whose hidden state is decoded into a joint interval distribution, refined through duration-aware coordinate correction, and converted into continuous boundaries.
The same interval representation is used to encode input temporal anchors, aligning video-side temporal evidence with output-side span prediction.
To reliably align the latent span representation with complete event intervals, we curate \textbf{90K-scale} grounded samples and human-verify \textbf{3K-scale} benchmark annotations.
Experiments across four VTG benchmarks show that \ours consistently outperforms endpoint prediction baselines, achieving an \textbf{average mIoU of 58.9}, with clear gains on short-duration and medium-duration events.

\end{abstract}
}

\begin{document}

\maketitle

\section{Introduction}
Multimodal language models have recently made rapid progress in video understanding~\cite{maaz2024videochatgpt, zhang2023videollama, lin2023videollava, zhang2024llavavideo, zhang2024longva, chen2024longvila, wang2025internvideo25, zhang2025videollama, wang2024qwen2vl, bai2025qwen25vl, bai2025qwen3vl}.
However, recognizing video content and grounding it precisely in time remain distinct capabilities.
Video temporal grounding (VTG) requires a model to localize the continuous temporal interval described by a natural-language query. 
At its core, the target is a coherent event interval rather than two timestamps considered separately.
Both supervision and evaluation are interval-level: a prediction is judged by its overlap with the complete event moment, whose temporal placement and duration jointly determine localization accuracy.
Yet current VLM-based formulations typically obtain this interval from two endpoint outputs, creating a mismatch between the task target and the model output.

Existing methods mainly differ in how temporal endpoints are represented.
One common direction represents time as discrete timestamp outputs, including textual timestamps~\cite{bai2025qwen3vl, zhang2024longva, deng2024seq2time, ren2024timechat, qian2024momentor, wu2024numpro, zhang2025vtimecot, wang2024timerefine, wang2025timer1, yue2025tempor0, li2025tempsampr1, dong2025videotgr1, mo2025tartvg, li2026videoopd} and symbolic time tokens~\cite{huang2024vtimellm, huang2024lita, guo2024vtgllm, wang2024groundedvideollm, li2025unitime}.
These formulations preserve the autoregressive language interface.
However, their token-level supervision does not reflect temporal distance on the video timeline: predictions close to and far from the ground-truth moment may both be treated as discrete token errors, despite implying substantially different localization quality.
Another direction decodes continuous start and end boundaries~\cite{jin2026should, zeng2025distime}, providing smoother temporal supervision without discrete time labels.
Nevertheless, it retains the same endpoint-centric prediction object.
Interval validity must be enforced through ordering constraints or post-processing, duration remains a difference between two coordinates, and interval-level similarity is not directly represented by the native output space.
Thus, discrete and continuous endpoint methods differ in how endpoints are encoded, but not in what is predicted: the event interval remains a derived object rather than the model primary output.

\begin{figure*}[!t]
  \centering
  \includegraphics[width=\textwidth]{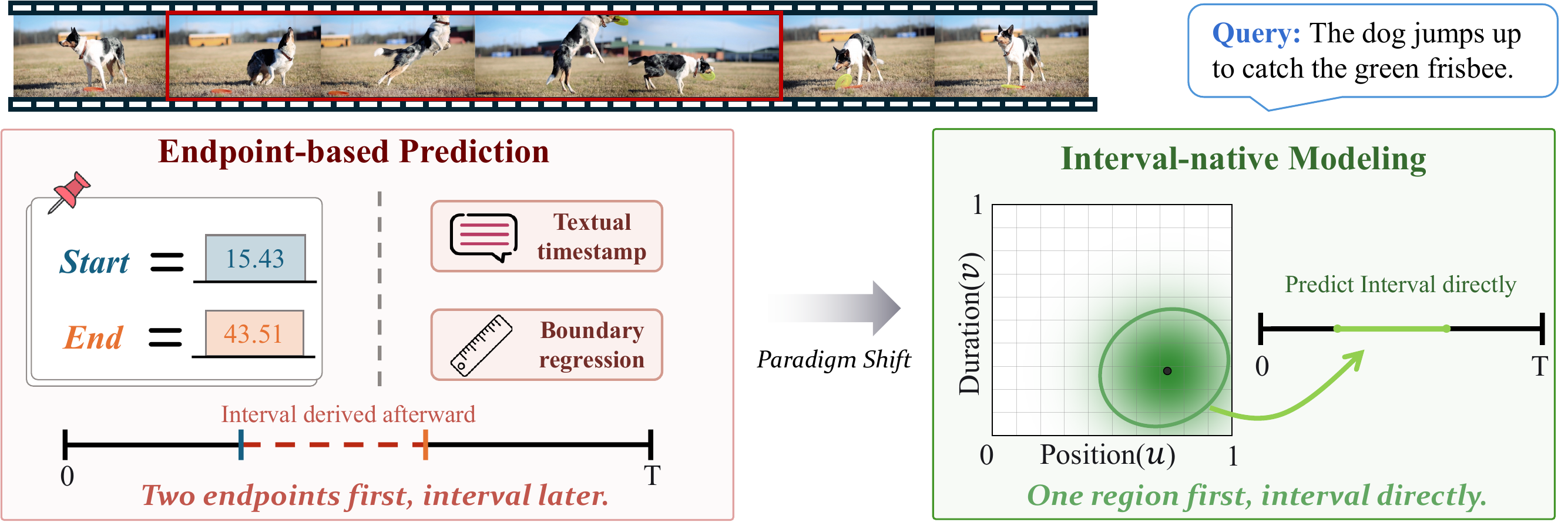}
  \caption{\textbf{From endpoint prediction to interval-native temporal grounding.}
  Existing methods represent a temporal moment through two endpoint outputs, encoded either as discrete timestamp tokens or continuous boundary coordinates, and assemble the interval afterward.
  TimePLE instead predicts a single joint distribution over valid temporal intervals, from which the final start and end boundaries are subsequently decoded.
  }
  \label{fig:teaser}
\end{figure*}

Motivated by this perspective, we propose \textbf{\ours}, which reformulates VTG from predicting two temporal endpoints to predicting a single joint distribution over valid temporal intervals.
We parameterize the space of normalized intervals using a canonical position--duration square.
For an interval $I=[s,e]$, its duration is represented by $v=e-s$, while
$u=s/(1-v)$ describes the feasible placement of an interval with that duration.
This mapping separates how long an event lasts from where an event of that length occurs on the timeline.
More importantly, every point in the canonical square corresponds to a valid temporal interval, and neighboring points represent geometrically similar spans.
The decoder can therefore assign probability directly to candidate event moments rather than to endpoint pairs.

\ours connects this interval-native output space to a VLM through a lightweight latent temporal interface.
Given a video and a query, the VLM generates a single \texttt{<|TIMESPAN|>} token whose hidden state parameterizes a joint distribution over the canonical interval square.
The expected canonical coordinate is refined through a duration-aware correction and then mapped back to continuous temporal boundaries.
On the input side, the temporal coverage of sampled visual units is represented by interval-valued anchors in the same canonical space and inserted through \texttt{<|TIMESTAMP|>} tokens.
This shared geometry aligns video-side temporal evidence with output-side interval prediction, while preserving the autoregressive VLM interface.


To reliably learn this latent interval interface from video-language semantics, we curate \textbf{90K-scale} grounded samples, each pairing a natural-language event description with the complete temporal interval it describes.
The interval is converted into a canonical target for supervising a single \texttt{<|TIMESPAN|>} representation, while \textbf{3K-scale} benchmark annotations are separately human-verified for reliable interval-level evaluation.
Across three VLM backbones and four VTG benchmarks, \ours consistently outperforms matched timestamp-based baselines.
On Qwen3-VL-8B, it achieves an \textbf{average mIoU of 58.9}, improving the matched Timestamp-SFT baseline by 4.1 points.
The gains are particularly clear on short-duration and medium-duration events, whose interval overlap is more sensitive to small localization errors.
Representation-level analyses further show that \ours draws more coherent evidence from the target moment and produces a more localized probability landscape over candidate intervals.

Our contributions are summarized as follows:
\begin{itemize}
    \item We reformulate VLM-based video temporal grounding from endpoint prediction to interval-native grounding.
    Instead of temporal endpoints prediction, the proposed formulation makes a joint distribution over valid intervals the primary prediction object.


    \item We introduce \textbf{\ours}, which parameterizes valid temporal intervals in a canonical position--duration square and decodes the hidden state of a single \texttt{<|TIMESPAN|>} token into an interval distribution.

    
    \item We curate \textbf{90K-scale} grounded samples that pair event descriptions with temporal intervals to support latent interval alignment, and human-verify \textbf{3K-scale} benchmark annotations for reliable interval-level evaluation.
    

    \item We evaluate \ours across four VTG benchmarks. 
    \ours achieves \textbf{58.9 average mIoU} across four VTG benchmarks and performs favorably against endpoints prediction methods, with clear gains on short-duration and medium-duration events.

\end{itemize}

\section{Related Work}

\subsection{VLM-based Temporal Video Grounding}
Recent VLM-based VTG methods make time accessible to multimodal language models through timestamp-aware video representations, relative or absolute time tokens, frame-number prompts, interleaved timestamp markers, and continuous time decoders~\cite{ren2024timechat, huang2024vtimellm, huang2024lita, guo2024vtgllm, wang2024groundedvideollm, qian2024momentor, deng2024seq2time, wu2024numpro, li2025unitime, zeng2025distime, jin2026should}.
These methods allow instruction-following VLMs to refer to temporal locations through the language interface.
Textual timestamps reuse the native vocabulary of language models, while temporal-token methods introduce symbolic time indices for finer temporal reference.
Continuous temporal decoders further replace discrete outputs with smoother
boundary-level supervision.
Despite these differences, existing formulations mainly change how temporal endpoints are represented rather than what is predicted.
In contrast, \ours makes the interval itself the primary prediction object by decoding a single joint distribution over a canonical space of valid temporal spans.

\subsection{Datasets and Benchmark Quality}

VTG datasets and benchmarks such as Charades-STA, ActivityNet-Captions, TACoS, DiDeMo, and QVHighlights have supported temporal grounding across diverse video domains, query styles, and evaluation settings~\cite{gao2017tall, krishna2017dense, regneri2013tacos, anne2017didemo, lei2021qvhighlights, soldan2022mad, grauman2022ego4d, zhou2018youcook2}.
However, VTG is sensitive to boundary quality, query ambiguity, and annotation completeness.
Recent work has highlighted data quality issues in existing VTG benchmarks and introduced re-annotated data for more reliable training and evaluation~\cite{zhang2025timelens}.
Our work focuses on annotation quality for interval-native learning.
We curate reliable caption-grounded training samples to align the latent span representation with complete event intervals, and separately human-verify benchmark annotations for reliable interval-level evaluation.

\begin{figure*}[t]
  \centering
  \includegraphics[width=\textwidth]{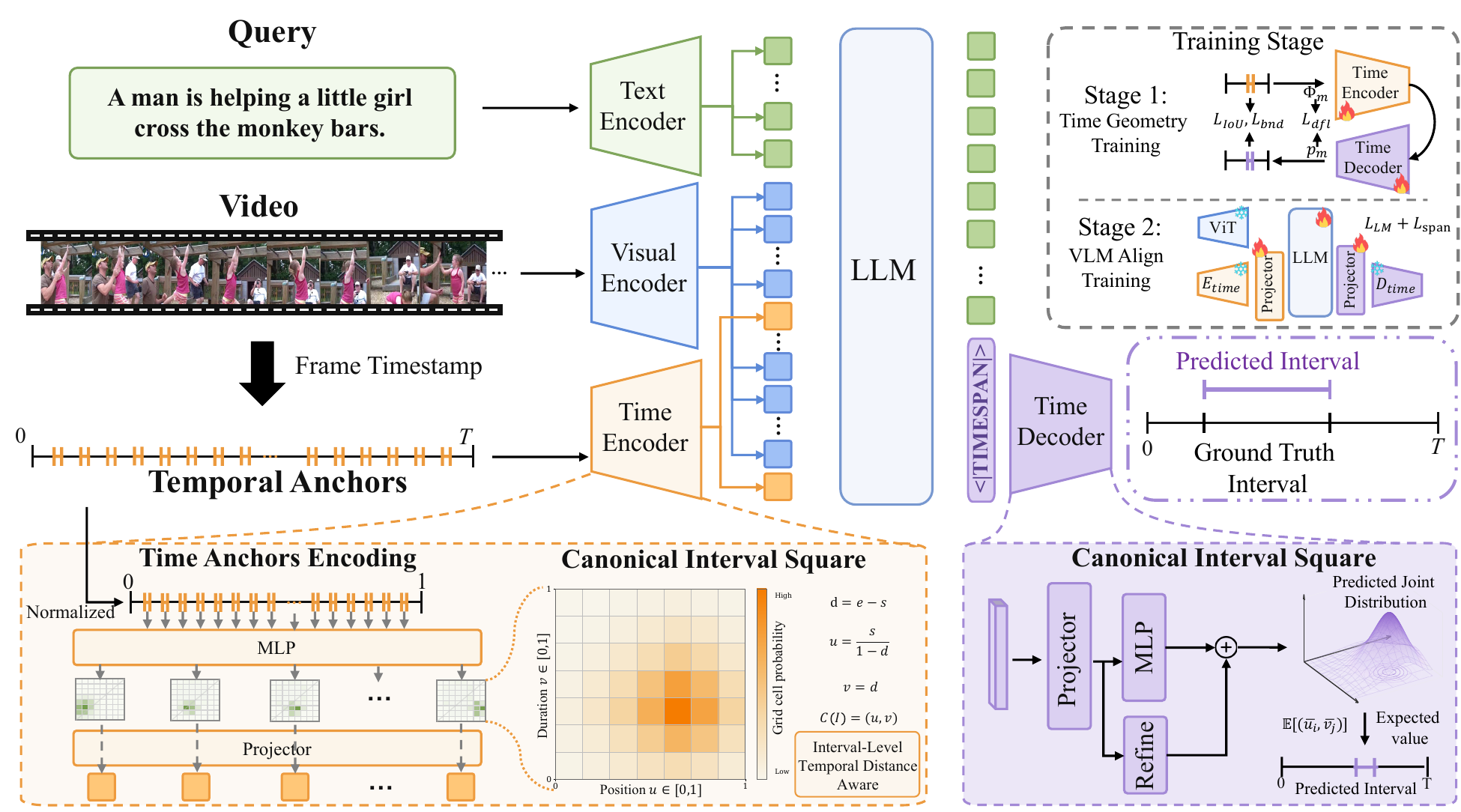}
  \caption{\textbf{Overview of TimePLE.} 
  Temporal anchors are encoded in a canonical position--duration space and inserted into the VLM through \texttt{<|TIMESTAMP|>} tokens.
  The hidden state of a generated \texttt{<|TIMESPAN|>} token is decoded into a joint distribution over valid intervals, refined, and mapped to continuous temporal boundaries.
  }
  \label{fig:framework}
\end{figure*}

\section{Method}

\subsection{Overview and Problem Formulation}
Given a video $V$ with duration $T$ and a natural-language query $q$, video temporal grounding aims to localize the temporal segment in $V$ that semantically corresponds to $q$. 
We denote the target segment in seconds as $\widetilde{I}=[t_s,t_e]$, where
$0 \leq t_s \leq t_e \leq T$.
For model prediction and supervision, we normalize the boundaries by the video duration and obtain $I=[s,e]$, where $s=t_s/T$, $e=t_e/T$, and
$0 \leq s \leq e \leq 1$.
TimePLE preserves the autoregressive interface of VLMs while
reformulating temporal localization as interval-distribution prediction.
As illustrated in Fig.~\ref{fig:framework}, the model generates a response containing \texttt{<|TIMESPAN|>}, whose hidden state parameterizes a latent interval distribution. 
TimePLE decodes the hidden state into a normalized interval $\hat{I}=[\hat{s},\hat{e}]$ in a canonical interval space and rescales it to seconds. 

\subsection{Canonical Interval Square}
\label{sec:3_2}
To parameterize a joint distribution over valid temporal intervals, \ours maps the feasible interval space to a canonical
position--duration square.
Let $d=e-s$ denote the normalized duration. 
For $d<1$, the feasible start position lies in $[0,1-d]$. We define the canonical coordinate as

\begin{equation}
\mathcal{C}(I)=(u,v),
\qquad
u=\frac{s}{1-d},
\qquad
v=d.
\label{eq:cis_forward}
\end{equation}
Here, $v$ directly represents the interval duration, while $u$ represents the relative placement of an interval with that duration within its feasible start range.
For the full-video interval, whose feasible start range collapses to $s=0$, we define $\mathcal{C}([0,1])=(0,1)$.
The inverse mapping reconstructs the normalized interval as
\begin{equation} 
\mathcal{C}^{-1}(u,v)
=
\left[
u(1-v),\;
u(1-v)+v
\right].
\label{eq:cis_inverse}
\end{equation}

We discretize the canonical square into an $N_u \times N_v$ grid  $\mathcal{G}=\{(i,j)\}$.
The grid centers and corresponding intervals are 
\begin{equation}
\bar{u}_i=\frac{i-1}{N_u-1},
\qquad
\bar{v}_j=\frac{j-1}{N_v-1},
\qquad
I_{ij}=\mathcal{C}^{-1}(\bar{u}_i,\bar{v}_j),
\label{eq:canonical_grid}
\end{equation}
where $1\leq i\leq N_u$ and $1\leq j\leq N_v$.

Given a continuous canonical coordinate $c=(u,v)$, we construct one-dimensional Gaussian distributions along the position and duration axes:

\begin{equation}
\begin{aligned}
p_i^{u}
&=
\operatorname{softmax}_{i}
\left(
-\frac{(\bar{u}_i-u)^2}{2\sigma_u^2}
\right),\\
p_j^{v}
&=
\operatorname{softmax}_{j}
\left(
-\frac{(\bar{v}_j-v)^2}{2\sigma_v^2}
\right).
\end{aligned}
\label{eq:axis_density}
\end{equation}
The canonical soft target is their outer product:
\begin{equation}
\Phi_{ij}(u,v)=p_i^{u}p_j^{v}.
\label{eq:grid_density}
\end{equation}

\subsection{TimePLE: Duration-Aware Interval Codec}
\label{sec:3_3}
Given the canonical mapping $\mathcal{C}$ and the soft density function $\Phi$, TimePLE uses a lightweight interval codec to connect temporal intervals with the hidden space of the VLM.
On the input side, temporal anchors are encoded as soft interval densities and inserted into the input token sequence.
On the output side, the hidden state of \texttt{<|TIMESPAN|>} is decoded into a distribution over the same canonical grid and refined through a duration-aware coordinate correction.

\paragraph{Distributional Span Decoding.}
Let $h_{\mathrm{span}}$ denote the hidden state of a generated \texttt{<|TIMESPAN|>} token.
We project this hidden state into the interval feature space and decode grid logits:
\begin{equation}
f_{\mathrm{span}}
=
A_{\mathrm{out}}(h_{\mathrm{span}}),
\quad
\boldsymbol{\ell}
=
D_{\mathrm{int}}(f_{\mathrm{span}})
\in
\mathbb{R}^{N_u\times N_v},
\label{eq:span_logits}
\end{equation}
where $A_{\mathrm{out}}$ is the output projection layer and
$D_{\mathrm{int}}$ is the interval decoder.

The logits define a joint span distribution over the canonical
grid:
\begin{equation}
p_{\theta}(i,j\mid h_{\mathrm{span}})
=
\operatorname{softmax}_{\mathcal{G}}
(\boldsymbol{\ell})_{ij}.
\label{eq:span_distribution}
\end{equation}
We obtain the initial continuous coordinate by taking the expectation over the grid:
\begin{equation}
(\hat{u}_0,\hat{v}_0)
=
\sum_{i=1}^{N_u}
\sum_{j=1}^{N_v}
p_{\theta}(i,j\mid h_{\mathrm{span}})
(\bar{u}_i,\bar{v}_j).
\label{eq:span_expectation}
\end{equation}
This expectation decoding preserves uncertainty in the predicted span distribution while producing a continuous interval coordinate.

\paragraph{Duration-Aware Coordinate Refinement.}
Although grid expectation provides a continuous prediction, its localization precision remains affected by grid discretization.
We therefore use the predicted duration coordinate $\hat{v}_0$ to determine the refinement strength:
\begin{equation}
g_0
=
\frac{1}{\kappa_{\mathrm{cap}}}
\operatorname{clip}
\left(
\frac{1}{\sqrt{\hat{v}_0+\epsilon_g}},
0,
\kappa_{\mathrm{cap}}
\right).
\label{eq:duration_gate}
\end{equation}
The gate assigns larger correction strength to shorter predicted intervals while keeping the refinement bounded.
We use $\epsilon_g=10^{-4}$ and $\kappa_{\mathrm{cap}}=10$.

Using the interval feature $f_{\mathrm{span}}$, the refinement module predicts a bounded coordinate offset:
\begin{equation}
\Delta \mathbf{c}
=
\alpha g_0
\tanh
\left(
R_{\mathrm{int}}(f_{\mathrm{span}})
\right),
\label{eq:refinement_offset}
\end{equation}
where $\Delta\mathbf{c}=(\Delta u,\Delta v)$, $R_{\mathrm{int}}$ denotes the refinement module, and $\alpha$ controls the maximum correction scale.
The refined coordinate and the final normalized interval are obtained by 
\begin{equation}
\hat{\mathbf{c}}
=
\operatorname{clamp}
\left(
(\hat{u}_0,\hat{v}_0)+\Delta\mathbf{c},
0,
1
\right),
\quad
\hat{I}
=
\mathcal{C}^{-1}(\hat{\mathbf{c}}).
\label{eq:refined_interval}
\end{equation}

\begin{figure*}[!t]
  \centering
  \includegraphics[width=\textwidth]{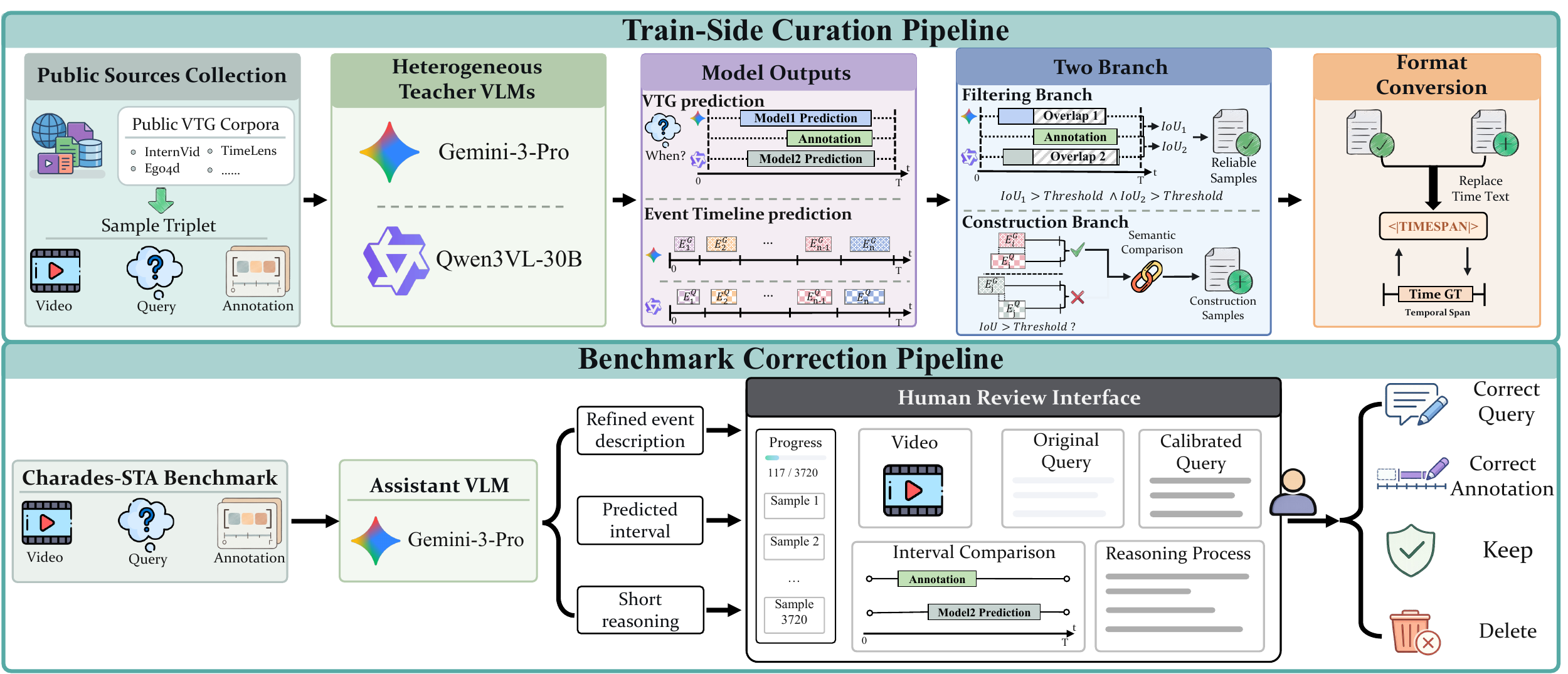}
  \caption{\textbf{Data curation and benchmark correction.}
  Heterogeneous teacher VLMs filter existing annotations by 
  cross-model agreement and construct new samples through temporal and semantic consensus. 
  Accepted moments are converted into TimePLE supervision, while model outputs serve only as auxiliary evidence for human benchmark correction.
  }
  \label{fig:data_pipeline}
\end{figure*}

\paragraph{Temporal Anchor Encoding.}
For the $r$-th sampled visual unit, we assign a normalized temporal anchor $a_r=[s_r^a,e_r^a]$ according to its temporal coverage in the video. 
The anchor is converted into a canonical soft density and projected into the VLM embedding space:
\begin{equation}
\begin{aligned}
\phi_r^a
&=
\Phi\!\left(\mathcal{C}(a_r)\right),\\
\mathbf{x}_{p_r}
&\leftarrow
A_{\mathrm{in}}
\left(
E_{\mathrm{int}}(\phi_r^a)
\right),
\end{aligned}
\label{eq:anchor_encoding}
\end{equation}
where $E_{\mathrm{int}}$ denotes the interval encoder, $A_{\mathrm{in}}$ is the input projection layer, and $p_r$ denotes the position of the corresponding \texttt{<|TIMESTAMP|>} token.
This operation represents the temporal coverage of each visual unit using the same canonical interval geometry employed for output prediction.

\subsection{Training Objectives}
\label{sec:3_4}
TimePLE is trained in two stages: interval geometry training and interval-supervised fine-tuning.
The first stage learns the canonical interval codec independently of the VLM, while the second stage aligns the generated \texttt{<|TIMESPAN|>} representation with the learned interval space.

\paragraph{Interval Geometry Training.}
We construct synthetic interval samples by uniformly sampling canonical coordinates within balanced grid cells.
Each coordinate $c_m=(u_m,v_m)$ is converted into a normalized interval $I_m=\mathcal{C}^{-1}(c_m)$, paired with a sampled video duration $T_m$, and represented by the target density $\phi_m=\Phi(c_m)$.
The interval encoder and decoder reconstruct the corresponding span distribution $p_m$ and refined interval $\hat{I}_m$.

For a target interval $I=[s,e]$, predicted interval $\hat{I}=[\hat{s},\hat{e}]$, target density $\phi$, predicted distribution $p$, and video duration $T$, we define the shared span objective as
\begin{equation}
\begin{aligned}
\mathcal{L}_{\mathrm{span}}
={}&
\lambda_{\mathrm{dfl}}
\operatorname{CE}(\phi,p)
+
\lambda_{\mathrm{iou}}
\left(
1-\operatorname{IoU}(\hat{I},I)
\right)\\
&+
\lambda_{\mathrm{bnd}}
\bar{w}(D)
\frac{
|\hat{s}-s|+|\hat{e}-e|
}{2},
\quad
D=T(e-s),
\end{aligned}
\label{eq:span_loss}
\end{equation}
where $\bar{w}(D)$ denotes the batch-normalized absolute-duration weight. 
The three terms respectively supervise the canonical interval distribution, interval overlap and boundary accuracy.

The interval geometry objective is
\begin{equation}
\mathcal{L}_{\mathrm{geo}}
=
\frac{1}{M}
\sum_{m=1}^{M}
\mathcal{L}_{\mathrm{span}}
\left(
p_m,\hat{I}_m;
\phi_m,I_m,T_m
\right).
\label{eq:geometry_loss}
\end{equation}

\paragraph{Supervised Fine-Tuning.}
In the second stage, each training sample contains a video $V$, query $q$, target response $x_{1:N}$, target interval $\widetilde{I}^{*}$, and video duration $T$. 
The target response contains one \texttt{<|TIMESPAN|>} token aligned with the continuous temporal label.
The standard autoregressive language modeling loss is
\begin{equation}
\mathcal{L}_{\mathrm{LM}}
=
-
\sum_{t=1}^{N}
\log
p_{\theta}
\left(
x_t
\mid
x_{<t},V,q
\right).
\label{eq:lm_loss}
\end{equation}

Let $h_{\mathrm{span}}$ denote the hidden state at the \texttt{<|TIMESPAN|>} position. 
The interval codec decodes it into a span distribution $p_{\theta}$ and a refined interval $\hat{I}$. 
The target segment $\widetilde{I}^{*}=[t_s^{*},t_e^{*}]$ is normalized by $T$ to obtain $I^{*}=[s^{*},e^{*}]$, and its distributional target is $\phi^{*}=\Phi(\mathcal{C}(I^{*}))$.
We apply the shared span objective in Eq.~\eqref{eq:span_loss} to obtain $\mathcal{L}_{\mathrm{span}}^{\mathrm{SFT}}$.
The final supervised fine-tuning objective is
\begin{equation}
\mathcal{L}_{\mathrm{SFT}}
=
\mathcal{L}_{\mathrm{LM}}
+
\lambda_{\mathrm{span}}
\mathcal{L}_{\mathrm{span}}^{\mathrm{SFT}}.
\label{eq:sft_loss}
\end{equation}

\subsection{Data Curation}
\label{sec:3_5}
We curate training data from public VTG sources~\cite{wang2024internvid, zhang2025timelens, grauman2022ego4d, geng2025longvale} using Qwen3-VL-30B~\cite{bai2025qwen3vl} and Gemini-3-Pro~\cite{deepmind2026gemini31pro} as heterogeneous teacher VLMs.
Existing annotations are retained only when both teachers agree with the labeled interval, while new grounded samples require temporal overlap and semantic consistency between teacher predictions.
Each accepted moment is converted into one \texttt{<|TIMESPAN|>} token paired with a continuous interval label. 
Separately, model predictions and calibrated queries are presented as auxiliary evidence in a human review interface for benchmark correction.
The pipeline produces 90K training samples and corrects 3K-scale noisy annotations. Additional details are provided in the Appendix.

\begin{table*}[!t]
\centering
\small
\setlength{\tabcolsep}{1.5pt}
\renewcommand{\arraystretch}{1.08}
\resizebox{\textwidth}{!}{
\begin{tabular}{l|c|ccc|cccc|ccc|cccc|c}
\toprule
\multirow{2}{*}{\textbf{Model}}
& \multirow{2}{*}{\textbf{Rep.}}
& \multicolumn{3}{c|}{\textbf{Charades-STA}}
& \multicolumn{4}{c|}{\textbf{ActivityNet-Captions}}
& \multicolumn{3}{c|}{\textbf{QVHighlights}}
& \multicolumn{4}{c|}{\textbf{Charades-TimePLE}}
& \multirow{2}{*}{\textbf{Avg.}} \\
\cmidrule(lr){3-5}
\cmidrule(lr){6-9}
\cmidrule(lr){10-12}
\cmidrule(lr){13-16}
& 
& \textbf{mIoU} & \textbf{$\mathrm{mIoU}_{S}$} & \textbf{$\mathrm{mIoU}_{M}$} 
& \textbf{mIoU} & \textbf{$\mathrm{mIoU}_{S}$} & \textbf{$\mathrm{mIoU}_{M}$} & \textbf{$\mathrm{mIoU}_{L}$}
& \textbf{mIoU} & \textbf{$\mathrm{mIoU}_{M}$} & \textbf{$\mathrm{mIoU}_{L}$}
& \textbf{mIoU} & \textbf{$\mathrm{mIoU}_{S}$} & \textbf{$\mathrm{mIoU}_{M}$} & \textbf{$\mathrm{mIoU}_{L}$}
& \\
\midrule

\multicolumn{17}{l}{\emph{Proprietary Models}} \\
GPT-4o~\cite{openai2024gpt4o} 
& T
& 22.5 & 19.8 & 31.9
& 19.2 & 19.5 & 18.7 & 19.3 
& 14.9 & 14.5 & 15.2 
& 28.7 & 24.9 & 39.2 & \textbf{74.7}
& 21.3 \\
Gemini-3.1-Pro~\cite{deepmind2026gemini31pro} 
& T
& 45.1 & 43.4 & 51.1
& \underline{46.3} & \textbf{46.1} & \underline{47.6} & 45.5 
& 59.7 & 58.6 & \underline{60.9} 
& 52.8 & 49.8 & 60.9 & 72.7
& \underline{51.0} \\

\midrule
\multicolumn{17}{l}{\emph{Open-Source Models}} \\
VTG-LLM~\cite{guo2024vtgllm} 
& S
& 34.2 & 33.2 & 37.7
& 16.7 & 19.3 & 22.8 & 10.3 
& 10.0 & 9.4 & 10.7 
& 33.9 & 32.2 & 39.0 & 34.7
& 23.7 \\
Grounded-VideoLLM~\cite{wang2024groundedvideollm} 
& S
& 42.2 & 39.9 & 50.2 
& 40.9 & 27.4 & 40.7 & 50.3 
& 46.9 & 50.6 & 42.7 
& 44.2 & 41.1 & 53.4 & 58.2
& 43.6 \\
NumPro~\cite{wu2024numpro} 
& T
& 24.7 & 22.9 & 30.9
& 16.4 & 13.8 & 21.2 & 14.7 
& 24.2 & 16.3 & 33.2 
& 25.9 & 22.4 & 35.8 & 66.3
& 22.8 \\
DisTime~\cite{zeng2025distime} 
& B
& \underline{49.4} & 46.1 & \underline{61.0}
& 46.1 & 27.1 & 44.4 & \textbf{62.1} 
& 50.7 & 47.2 & 54.7 
& 53.4 & 48.9 & \underline{66.9} & \underline{73.4}
& 49.9 \\
Time-R1-7B~\cite{wang2025timer1} 
& T
& 28.5 & 28.6 & 27.9
& 16.3 & 20.1 & 20.4 & 10.6 
& 17.8 & 15.4 & 20.7 
& 28.4 & 27.8 & 30.3 & 28.5
& 22.8 \\
TimeLens-8B~\cite{zhang2025timelens} 
& T
& 42.6 & 40.7 & 48.9
& 44.0 & \underline{42.2} & 46.7 & 43.3 
& \underline{61.7} & \underline{62.2} & 59.3 
& 51.6 & 48.9 & 59.0 & 71.7
& 50.0 \\
InternVL3.5-8B~\cite{wang2025internvl35} 
& T
& 21.9 & 18.6 & 33.3 
& 27.9 & 25.2 & 27.2 & 30.3 
& 47.2 & 46.4 & 48.3 
& 28.1 & 23.9 & 39.7 & 66.4
& 31.3 \\
\rowcolor{gray!10}
Qwen3VL-8B~\cite{bai2025qwen3vl} 
& T
& 47.9 & \underline{46.4} & 53.1
& 42.3 & 39.7 & 43.8 & 42.8 
& 46.0 & 44.5 & 47.8 
& \underline{55.8} & \underline{53.9} & 61.2 & 69.8
& 48.0 \\
\rowcolor{cyan!10}
\textbf{TimePLE-8B}
& \textbf{I}
& \textbf{57.2} & \textbf{55.5} & \textbf{62.9}
& \textbf{49.6} & 37.2 & \textbf{50.1} & \underline{57.6}
& \textbf{65.3} & \textbf{65.9} & \textbf{64.2}
& \textbf{63.4} & \textbf{60.8} & \textbf{71.1} & 70.9
& \textbf{58.9} \\

\bottomrule
\end{tabular}
}
\caption{\textbf{Main results on VTG benchmarks.}
We report overall and duration-stratified mIoU.
Rep. denotes textual timestamps (T), symbolic time tokens
(S), continuous boundary decoding (B), and interval-native
decoding (I). Avg. is the mean overall mIoU across the four
benchmarks.
}
\label{tab:main_results}

\end{table*}
\section{Experiment}

\subsection{Experimental Setup}
We evaluate on Charades-STA, QVHighlights, ActivityNet-Captions, and the Charades-TimePLE benchmark.
TimePLE is built on Qwen3-VL-8B and uses 2-FPS video sampling with at most 200 frames and 64 visual tokens per frame.
We report overall mIoU and duration-stratified mIoU over Short $(0,10]$s, Medium $(10,30]$s, and Long $(30,\infty)$s moments.
More setup details are provided in the Appendix.

\subsection{Main Results on VTG Benchmarks}
\label{sec:main_results}

Table~\ref{tab:main_results} compares four temporal-output paradigms for VLM-based VTG.
TimePLE achieves \textbf{the best average performance of 58.9 mIoU} and improves Qwen3-VL-8B from 48.0 to 58.9.
The gains are strongest on short and medium moments, where small temporal errors cause larger degradation in interval overlap, while TimePLE remains competitive on long moments.
These results show that duration-conditioned interval prediction improves fine-grained localization without sacrificing coarse temporal grounding.

\subsection{Scaling Across VLM Backbones}
\newcommand{\scalingdelta}[2]{#1\,{\scriptsize #2}}
\newcommand{\timeplecell}[1]{\cellcolor{cyan!8}#1}

\begin{table}[t]
\centering
\caption{\textbf{Scaling across VLM backbones.}
TimePLE is compared with Timestamp-SFT baselines under the same
training data and video input configuration.
All entries are mIoU. C-STA, A-Net, QVH, and C-TPLE denote
Charades-STA, ActivityNet-Captions, QVHighlights, and Charades-TimePLE,
respectively.
Parenthesized numbers report mIoU gains over the matched
Timestamp-SFT baseline.
The Avg. gain is averaged across the four benchmarks.
}
\label{tab:backbone_scaling}
\footnotesize
\setlength{\tabcolsep}{1.2pt}
\renewcommand{\arraystretch}{1.10}

\begin{tabular}{@{}llccccc@{}}
\toprule
Backbone & Method
& C-STA & A-Net & QVH & C-TPLE & Avg. \\
\midrule

\multirow{2}{*}{LLaVA-OV-0.5B}
& Timestamp-SFT
& 23.8 & 10.8 & 6.5 & 24.7 & 16.5 \\
& \timeplecell{TimePLE}
& \timeplecell{\scalingdelta{\textbf{40.2}}{\gain{16.4}}}
& \timeplecell{\scalingdelta{\textbf{33.2}}{\gain{22.4}}}
& \timeplecell{\scalingdelta{\textbf{32.3}}{\gain{25.8}}}
& \timeplecell{\scalingdelta{\textbf{42.3}}{\gain{17.6}}}
& \timeplecell{\scalingdelta{\textbf{37.0}}{\gain{20.6}}} \\
\midrule

\multirow{2}{*}{Qwen2.5-VL-3B}
& Timestamp-SFT
& 51.2 & 41.6 & 59.3 & 57.8 & 52.5 \\
& \timeplecell{TimePLE}
& \timeplecell{\scalingdelta{\textbf{52.3}}{\gain{1.1}}}
& \timeplecell{\scalingdelta{\textbf{43.3}}{\gain{1.7}}}
& \timeplecell{\scalingdelta{\textbf{59.4}}{\gain{0.1}}}
& \timeplecell{\scalingdelta{\textbf{58.5}}{\gain{0.7}}}
& \timeplecell{\scalingdelta{\textbf{53.4}}{\gain{0.9}}} \\
\midrule

\multirow{2}{*}{Qwen3-VL-8B}
& Timestamp-SFT
& 51.6 & 47.3 & 63.6 & 56.7 & 54.8 \\
& \timeplecell{TimePLE}
& \timeplecell{\scalingdelta{\textbf{57.2}}{\gain{5.6}}}
& \timeplecell{\scalingdelta{\textbf{49.6}}{\gain{2.3}}}
& \timeplecell{\scalingdelta{\textbf{65.3}}{\gain{1.7}}}
& \timeplecell{\scalingdelta{\textbf{63.4}}{\gain{6.7}}}
& \timeplecell{\scalingdelta{\textbf{58.9}}{\gain{4.1}}} \\
\bottomrule
\end{tabular}

\vspace{-3.0pt}
\end{table}

We further apply the same interval-native
formulation to LLaVA-OV-0.5B~\cite{li2024llava} and Qwen2.5-VL-3B~\cite{bai2025qwen25vl}.
For each backbone, TimePLE is compared with a matched Timestamp-SFT baseline using the same training data and video input configuration, with only the temporal prediction interface changed.
Table~\ref{tab:backbone_scaling} shows higher mIoU for TimePLE in all twelve backbone--benchmark comparisons.
The average improvements are 20.6 points on LLaVA-OV-0.5B, 0.9 points on Qwen2.5-VL-3B, and 4.1 points on Qwen3-VL-8B.
The consistent gains across model families and scales support the benefit of replacing endpoint prediction with interval-distribution prediction.
The smaller improvements on the Qwen-VL backbones may result from their larger-scale pretraining and stronger language priors, which make Timestamp-SFT a more competitive baseline.

\begin{figure}[t]
    \centering
    \includegraphics[width=\textwidth]{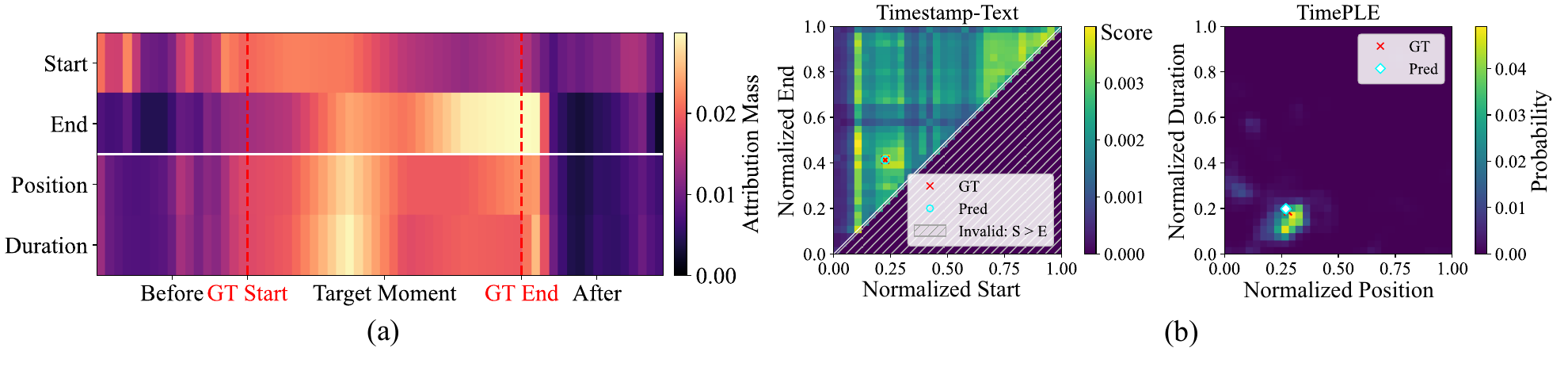}
    \caption{\textbf{Representation-level analysis.}
    (a) Gradient $\times$ Input attribution for Timestamp-Text start--end
    and TimePLE position--duration targets. Dashed lines denote ground-truth
    boundaries.
    (b) Timestamp-Text likelihood reconstructed over valid start--end
    candidates versus TimePLE's native distribution over the canonical
    interval square. Crosses and diamonds denote ground truth and prediction.}
    \label{fig:representation_analysis}
\end{figure}

\subsection{Representation-Level Analysis}
Beyond benchmark performance, we examine how endpoint and interval-distribution prediction differ in temporal representation, considering both supporting visual evidence and probability organization over candidate intervals.
More implementation details are provided in the Appendix.

\subsubsection{Temporal evidence attribution}
We first examine how the two temporal interfaces draw evidence from the video using Gradient $\times$ Input attribution on visual embeddings.
Timestamp-Text uses the teacher-forced likelihoods of the ground-truth start and end tokens, while TimePLE uses the soft marginal likelihoods of the ground-truth canonical position and duration.
As shown in Fig.~\ref{fig:representation_analysis}(a), endpoint prediction relies on temporally separated cues associated with the two boundaries.
In contrast, position and duration receive coherent support throughout the target moment, suggesting that TimePLE forms its prediction from the event segment as a whole rather than composing it from two endpoint decisions.

\subsubsection{Interval likelihood landscape}
We further visualize how temporal uncertainty is organized in the output space.
For Timestamp-Text, we enumerate valid boundary candidates and normalize their teacher-forced numeric-token likelihoods to reconstruct an interval landscape.
TimePLE directly provides a joint probability distribution over the canonical interval square.
In Fig.~\ref{fig:representation_analysis}(b), the reconstructed timestamp landscape is diffuse and fragmented over the triangular valid region, whereas TimePLE concentrates probability around the target in a fully valid interval space.
This comparison illustrates that interval-native decoding provides a more structured and localized distribution over candidate moments.

\newcommand{\stagedelta}[2]{#1\,{\scriptsize #2}}

\begin{table}[t]
\centering
\caption{
\textbf{Ablation of progressive training stages.}
All entries are mIoU. C-STA, A-Net, QVH, and C-TPLE denote
Charades-STA, ActivityNet-Captions, QVHighlights, and Charades-TimePLE,
respectively.
Parenthesized numbers report mIoU changes relative to the preceding training setting.
The Avg. gain is averaged across the four benchmarks.
}
\label{tab:progressive_training}
\footnotesize
\setlength{\tabcolsep}{1.2pt}
\renewcommand{\arraystretch}{1.10}
\begin{tabular}{@{}lccccc@{}}
\toprule
Training setting
& C-STA & A-Net & QVH & C-TPLE & Avg. \\
\midrule
Baseline
& 47.9 & 42.3 & 46.0 & 55.8 & 48.0 \\

\qquad + SFT
& \stagedelta{51.6}{\gain{3.7}}
& \stagedelta{47.3}{\gain{5.0}}
& \stagedelta{63.6}{\gain{17.6}}
& \stagedelta{56.7}{\gain{0.9}}
& \stagedelta{54.8}{\gain{6.8}} \\

\rowcolor{cyan!8}
TimePLE (SFT)
& \stagedelta{\textbf{57.2}}{\gain{5.6}}
& \stagedelta{\textbf{49.6}}{\gain{2.3}}
& \stagedelta{\textbf{65.3}}{\gain{1.7}}
& \stagedelta{\textbf{63.4}}{\gain{6.7}}
& \stagedelta{\textbf{58.9}}{\gain{4.1}} \\

\rowcolor{gray!8}
TimePLE (GRPO)
& \stagedelta{57.2}{\same{0.0}}
& \stagedelta{49.6}{\same{0.0}}
& \stagedelta{65.1}{\dropv{0.2}}
& \stagedelta{63.3}{\dropv{0.1}}
& \stagedelta{58.8}{\dropv{0.1}} \\
\bottomrule
\end{tabular}
\vspace{-3.0pt}
\end{table}

\begin{figure*}[!t]
    \centering
    \includegraphics[width=\textwidth]{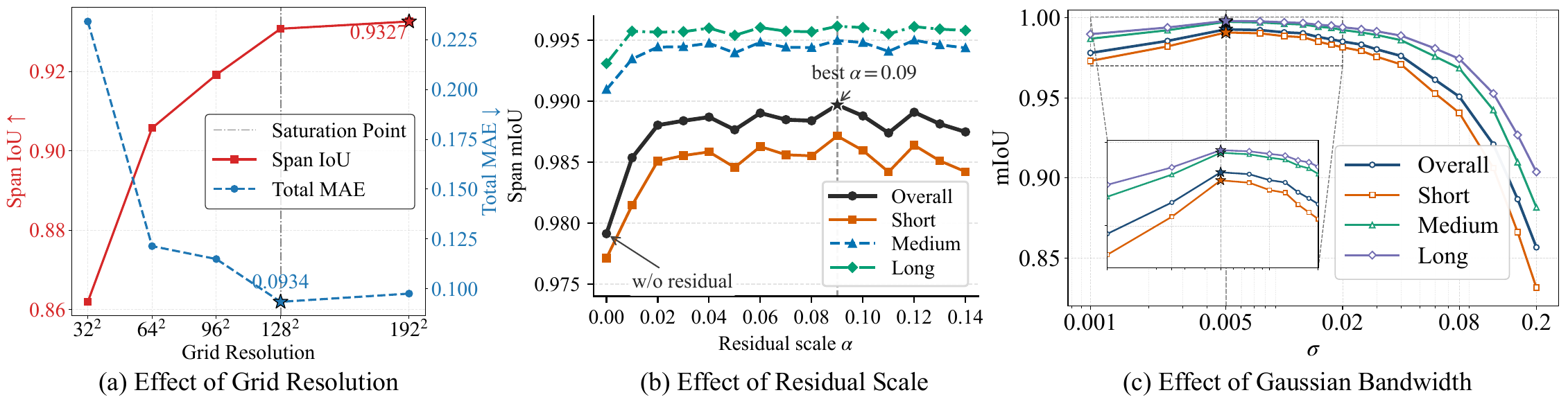}
    \caption{\textbf{Interval-space reconstruction analysis.}
    From left to right, we study the canonical grid resolution, residual refinement scale \(\alpha\), and Gaussian bandwidth \(\sigma\).
    Reconstruction largely saturates at \(128^2\), benefits from a small residual correction with \(\alpha=0.09\), and performs best under moderately local supervision with \(\sigma=0.005\).
    }
    \label{fig:interval_reconstruction}
\end{figure*}

\subsection{Ablation Studies}
\label{sec:ablation}
\subsubsection{Interval-Space Reconstruction Analysis}
We analyze whether the canonical interval space can support accurate continuous interval reconstruction under different discretization and smoothing settings.
\par
\noindent
\textbf{Grid resolution.}
As shown in Fig.~\ref{fig:interval_reconstruction} (a), reconstruction quality improves substantially as the grid increases from $32^2$ to $128^2$.
Increasing the resolution to $192^2$ yields only a marginal span-IoU gain, while the total MAE slightly increases and the number of grid support points grows by $2.25\times$.
We use $128^2$ as the default resolution, which provides a favorable balance between reconstruction accuracy and decoding complexity.
\par
\noindent
\textbf{Residual refinement.}
Setting $\alpha=0$ corresponds to direct grid-expectation decoding without coordinate refinement.
As shown in Fig.~\ref{fig:interval_reconstruction} (b), nonzero residual scales consistently improve reconstruction, with the largest gain observed for short intervals, which are more sensitive to small coordinate errors.
Performance remains stable across a broad range of scales and we set $\alpha=0.09$ as default.
\par
\noindent
\textbf{Gaussian bandwidth.}
We further analyze the Gaussian bandwidth $\sigma$ used for soft interval-grid supervision.
Figure~\ref{fig:interval_reconstruction} (c) shows that reconstruction improves as $\sigma$ increases from an overly sharp target to a moderate bandwidth.
Larger bandwidths consistently degrade reconstruction, especially for short intervals, because excessive smoothing reduces the distinction between neighboring spans.
We therefore set $\sigma=0.005$ as default.

\begin{findingbox}
Accurate interval reconstruction requires both sufficient coverage and localized distributional supervision.
Residual refinement complements grid-based decoding by correcting the remaining continuous localization error, particularly for short intervals.
\end{findingbox}

\subsubsection{Effect of Interval-Supervised Training}

Table~\ref{tab:progressive_training} separates ordinary task adaptation from interval-space supervision.
Timestamp-SFT raises the average mIoU from 48.0 to 54.8, confirming the benefit of supervised adaptation to the VTG response format.
TimePLE-SFT further improves the average to 58.9 and yields consistent gains on all four benchmarks, isolating the effect of interval-level supervision beyond ordinary SFT.
In contrast, GRPO produces no additional improvement, with the average mIoU slightly changing from 58.9 to 58.8.
A more detailed analysis of post-SFT optimization is provided in the Appendix.

\begin{findingbox}

The main gain of TimePLE comes from directly aligning the
span-token representation with the canonical interval space.
Once this alignment is learned through interval-supervised SFT,
outcome-level GRPO provides little additional benefit.
\end{findingbox}

\section{Conclusion}
We present \ours, an interval-native framework for VLM-based video temporal grounding.
Rather than generating timestamps or boundary pairs, \ours decodes a generated span token into a distribution over duration-conditioned event intervals.
This representation preserves the autoregressive VLM interface while making duration, placement, and interval-level supervision explicit in the prediction space.
Experiments across four VTG benchmarks show consistent gains, especially on short and medium-duration events.

\clearpage
\bibliographystyle{plainnat}
\bibliography{references}

\clearpage
\beginappendix
\noindent\textbf{Appendix Contents}
\vspace{0.4em}

\begingroup
\small
\setlength{\parindent}{0pt}
\setlength{\parskip}{2pt}

\hyperref[app:implementation_details]{%
A\quad Implementation Details}\\
\hspace*{1.5em}%
\hyperref[app:vlm_temporal_token_interface]{%
A.1\quad VLM Backbone and Temporal Token Interface}\\
\hspace*{1.5em}%
\hyperref[app:interval_codec_configuration]{%
A.2\quad Interval Codec Configuration}\\

\hyperref[app:canonical_interval_space]{%
B\quad Additional Details of the Canonical Interval Space}\\
\hspace*{1.5em}%
\hyperref[app:canonical_parameterization]{%
B.1\quad Canonical Position-Duration Parameterization}\\
\hspace*{1.5em}%
\hyperref[app:cis_vs_boundary_space]{%
B.2\quad Theoretical Analysis of the Canonical Interval Square}\\

\hyperref[app:representation_analysis]{%
C\quad Additional Details of Representation-Level Analysis}\\
\hspace*{1.5em}%
\hyperref[app:temporal_evidence_attribution]{%
C.1\quad Temporal Evidence Attribution}\\
\hspace*{1.5em}%
\hyperref[app:interval_likelihood_landscape]{%
C.2\quad Interval Likelihood Landscape}\\

\hyperref[app:data_curation_benchmark_correction]{%
D\quad Data Curation and Benchmark Correction Details}\\
\hspace*{1.5em}%
\hyperref[app:data_pipeline_overview]{%
D.1\quad Data Pipeline Overview}\\
\hspace*{1.5em}%
\hyperref[app:training_data_curation]{%
D.2\quad Training Data Curation and Construction}\\
\hspace*{1.5em}%
\hyperref[app:benchmark_correction]{%
D.3\quad Benchmark Correction with Human Review}\\
\hspace*{1.5em}%
\hyperref[app:data_statistics]{%
D.4\quad Data Statistics}\\

\hyperref[app:evaluation_metric_design]{%
E\quad Evaluation Metric Design}\\
\hspace*{1.5em}%
\hyperref[app:standard_temporal_grounding_metrics]{%
E.1\quad Standard Temporal Grounding Metrics}\\
\hspace*{1.5em}%
\hyperref[app:duration_stratified_metric]{%
E.2\quad Duration-Stratified mIoU and Boundary-Tolerance Analysis}\\

\hyperref[app:progressive_interval_analysis]{%
F\quad Additional Exploration of Progressive Interval Optimization}\\
\hspace*{1.5em}%
\hyperref[app:post_sft_motivation]{%
F.1\quad Motivation and Setup}\\
\hspace*{1.5em}%
\hyperref[app:post_sft_algorithms]{%
F.2\quad Explored Post-SFT Training Algorithms}\\
\hspace*{1.5em}%
\hyperref[app:post_sft_results]{%
F.3\quad Experimental Results}\\
\hspace*{1.5em}%
\hyperref[app:post_sft_discussion]{%
F.4\quad Mechanistic Analysis of Post-SFT Optimization}

\endgroup

\vspace{0.8em}
\hrule
\vspace{1.0em}
\section{Implementation Details}
\label{app:implementation_details}
We provide the implementation details of TimePLE but are not expanded in the main paper.
\subsection{VLM Backbone and Temporal Token Interface}
\label{app:vlm_temporal_token_interface}
TimePLE is implemented on top of Qwen3-VL-8B-Instruct. To support interval-aware temporal grounding, we add two temporal special tokens, \texttt{<|TIMESTAMP|>} and \texttt{<|TIMESPAN|>}. The former is used for input-side temporal anchoring, while the latter provides the output-side latent interface for continuous interval decoding.
The newly introduced temporal-token embeddings are initialized from the mean and covariance statistics of the existing token embeddings. This provides stable initial representations for the temporal tokens without perturbing the original vocabulary embedding space.

\subsection{Interval Codec Configuration}
\label{app:interval_codec_configuration}
The interval encoder, interval decoder, and CIS span projector are implemented
as lightweight MLP modules. The encoder maps the input interval-grid
representation into the 4096-dimensional hidden space of the base VLM, the
decoder maps the 4096-dimensional interval feature back to the canonical
interval-grid prediction space, and the CIS span projector transforms the
hidden state associated with \texttt{<|TIMESPAN|>} before interval decoding.

\section{Additional Details of the Canonical Interval Space}
\label{app:canonical_interval_space}

\subsection{Canonical Position-Duration Parameterization}
\label{app:canonical_parameterization}
The canonical position-duration parameterization follows directly from the
feasible geometry of temporal intervals.
For a fixed duration, the start
position is only valid within the duration-conditioned range $[0,1-d]$ rather
than the full unit interval.
Therefore, normalizing the start position by this feasible range yields a duration-conditioned placement coordinate, while keeping the duration itself as an explicit coordinate.
This design represents a temporal
moment by where an interval of a given length can be placed, rather than by two
raw boundary values whose feasibility must be handled implicitly.

\subsection{Theoretical Analysis of the Canonical Interval Square}
\label{app:cis_vs_boundary_space}
Building on the parameterization above, the key theoretical property of the
canonical interval square is the alignment between the decoder output support
and the feasible interval space. Since each point in the canonical square
corresponds to a valid temporal interval, a dense grid decoder can be interpreted
directly as a distribution over valid spans. This avoids assigning probability
mass to invalid boundary configurations and removes the need for boundary
ordering, masking, or post-hoc interval repair.
This prediction geometry is more structured than directly modeling raw start and end boundaries jointly. In a boundary-coordinate space, interval validity, duration, and placement are entangled in two boundary values, and the rectangular output space of a neural decoder does not naturally coincide with the feasible set of ordered intervals.
In contrast, the canonical interval square explicitly separates temporal extent from feasible placement, allowing the model to express uncertainty over ``how long the moment lasts'' and ``where a moment of that length is located'' on two semantically meaningful axes. This provides a cleaner and more interval-native prediction domain for duration-aware temporal grounding.

\section{Additional Details of Representation-Level Analysis}
\label{app:representation_analysis}
We provide implementation details and additional qualitative examples for the representation-level analysis.
The analysis examines the two temporal interfaces from complementary perspectives: the visual evidence supporting their temporal targets and the organization of probability mass over candidate intervals.

\begin{figure*}[!t]
    \centering
    \includegraphics[width=\textwidth]{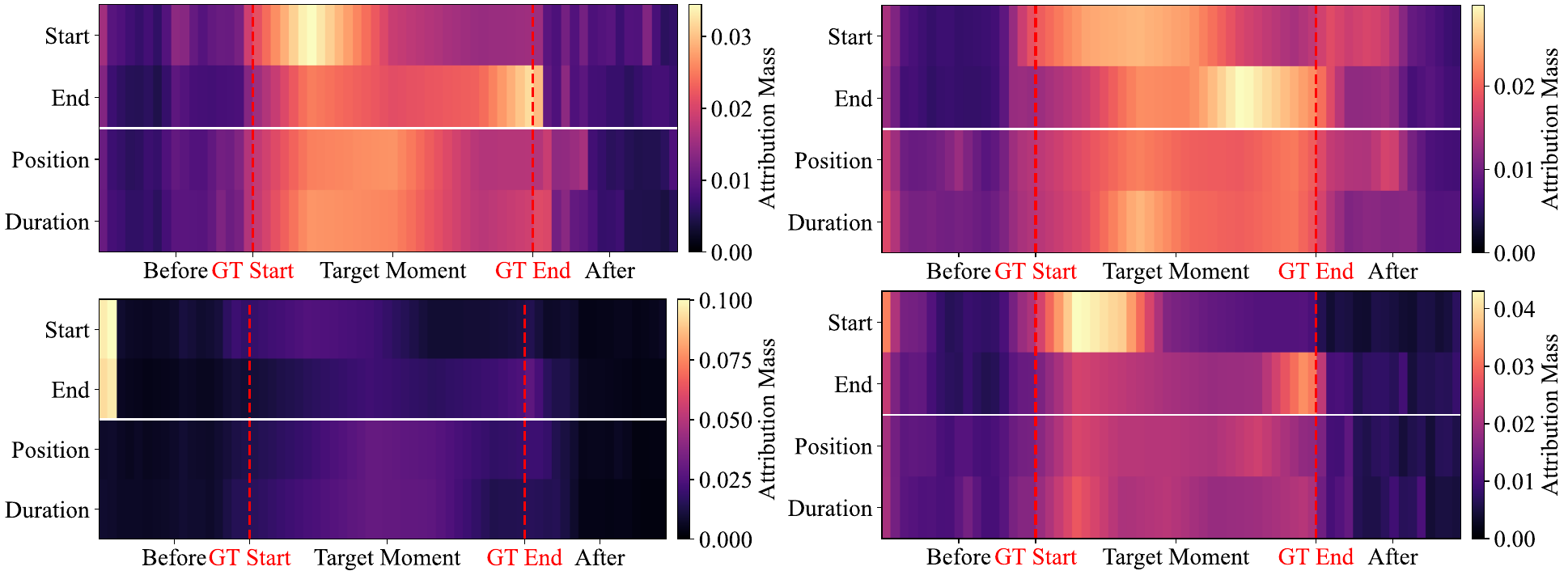}
    \caption{
    \textbf{Additional temporal evidence attribution cases.}
    Timestamp-Text attribution is computed from the
    teacher-forced likelihoods of the ground-truth start and end
    timestamp tokens, while TimePLE attribution is computed from
    the soft marginal likelihoods of the ground-truth canonical
    position and duration. Dashed lines denote the ground-truth
    temporal boundaries.
    }
    \label{fig:additional_attribution_cases}
\end{figure*}

\begin{figure*}[t]
    \centering
    \includegraphics[width=\textwidth]
    {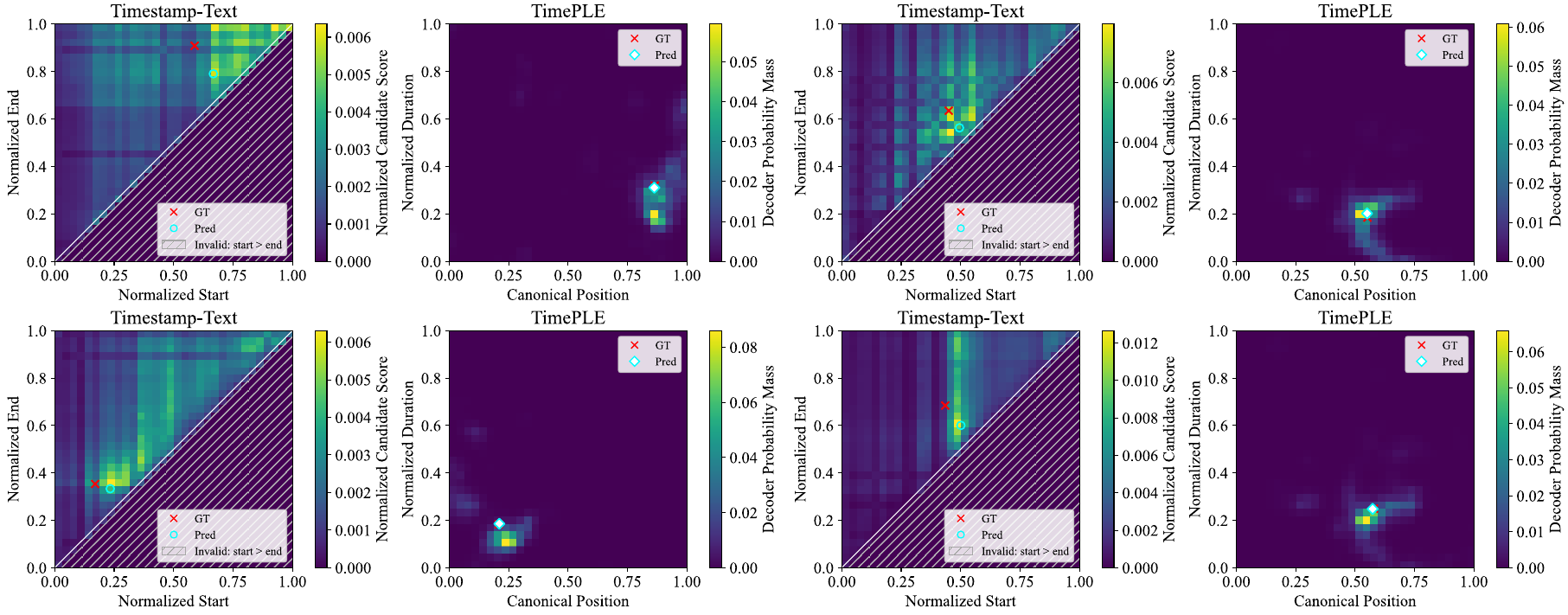}
    \caption{
    \textbf{Additional interval likelihood landscapes.}
    Timestamp-Text landscapes are reconstructed by enumerating
    528 valid start--end candidates and normalizing their
    teacher-forced numerical-token likelihoods. TimePLE directly
    outputs a joint probability distribution over the canonical
    interval square. Crosses and diamonds denote the ground truth
    and model prediction, respectively.
    }
    \label{fig:additional_landscape_cases}
\end{figure*}

\subsection{Temporal Evidence Attribution}
\label{app:temporal_evidence_attribution}
We use Gradient $\times$ Input attribution to measure how each temporal portion of the video supports the target temporal prediction.
We register a hook after the visual projection layer and extract the visual-token embeddings $\mathbf{E}\in\mathbb{R}^{N_{\mathrm{visual}}\times d}$.
Model parameters are frozen, while gradients are retained for $\mathbf{E}$.

For Timestamp-Text, the model is evaluated with teacher forcing using the ground-truth response.
Let $\mathcal{I}_{s}$ and $\mathcal{I}_{e}$ denote the numeric-token positions of the ground-truth start and end timestamps.
Their attribution targets are the mean token log-likelihoods
\begin{equation}
\begin{aligned}
y_s &=
\frac{1}{|\mathcal{I}_s|}
\sum_{k\in\mathcal{I}_s}
\log p_\theta(z_k\mid z_{<k},V,Q), \\
y_e &=
\frac{1}{|\mathcal{I}_e|}
\sum_{k\in\mathcal{I}_e}
\log p_\theta(z_k\mid z_{<k},V,Q).
\end{aligned}
\end{equation}
Averaging avoids attribution-scale differences caused by different numbers of numerical tokens.

For TimePLE, the ground-truth interval is mapped to its canonical position and duration.
From the predicted joint distribution $P(u,v)$, we obtain the corresponding marginal distributions and evaluate them using Gaussian soft targets $q_u$ and $q_v$ with $\sigma=0.015$:
\begin{equation}
\begin{aligned}
y_u &= \sum_i q_u(i)
\log \sum_j P(u_i,v_j), \\
y_v &= \sum_j q_v(j)
\log \sum_i P(u_i,v_j).
\end{aligned}
\end{equation}

For each scalar target \(y\), the attribution of visual token \(n\) is
\begin{equation}
a_n =
\sum_{c=1}^{d}
\left|
E_{n,c}
\frac{\partial y}{\partial E_{n,c}}
\right|.
\end{equation}
The attribution values of spatial tokens belonging to the same temporal position are averaged, producing a one-dimensional attribution sequence over the video timeline.
Larger values indicate that the corresponding visual content is more sensitive to the specified temporal target. Additional examples are shown in Fig.~\ref{fig:additional_attribution_cases}.

\subsection{Interval Likelihood Landscape}
\label{app:interval_likelihood_landscape}
Timestamp-Text does not natively produce a probability distribution over complete temporal intervals.
We uniformly discretize the video duration into 32 time points and enumerate the 528 valid start--end candidates.
Each candidate is formatted as a textual timestamp response and evaluated through teacher-forced numerical-token likelihood:
\begin{equation}
\begin{aligned}
S(i,j)
&=
\frac{1}{N_{ij}}
\sum_{k\in\mathcal{I}_{ij}}
\log p_\theta(z_k\mid z_{<k},V,Q), \\
P_{\mathrm{text}}(i,j)
&=
\frac{\exp S(i,j)}
{\sum_{a\leq b}\exp S(a,b)},
\qquad i\leq j.
\end{aligned}
\end{equation}
The region \(i>j\) contains invalid intervals and is masked in the visualization. Averaging over numerical tokens prevents candidate scores from being affected by timestamp-token length.

TimePLE instead directly outputs joint logits \(L\in\mathbb{R}^{128\times128}\) over the canonical position--duration square.
Its native interval distribution is
\begin{equation}
P_{\mathrm{TimePLE}}(u,v)
=
\frac{\exp L(u,v)}
{\sum_{a,b}\exp L(a,b)}.
\end{equation}
Every support point corresponds to a valid temporal interval.
For visualization, the distribution is reduced to $32\times32$ by summing each non-overlapping $4\times4$ block, which preserves the total probability mass.
Additional cases are provided in Fig.~\ref{fig:additional_landscape_cases}.

\section{Data Curation and Benchmark Correction Details}
\label{app:data_curation_benchmark_correction}
\subsection{Data Pipeline Overview}
\label{app:data_pipeline_overview}

\begin{figure*}[!t]
    \centering
    \includegraphics[width=\textwidth]{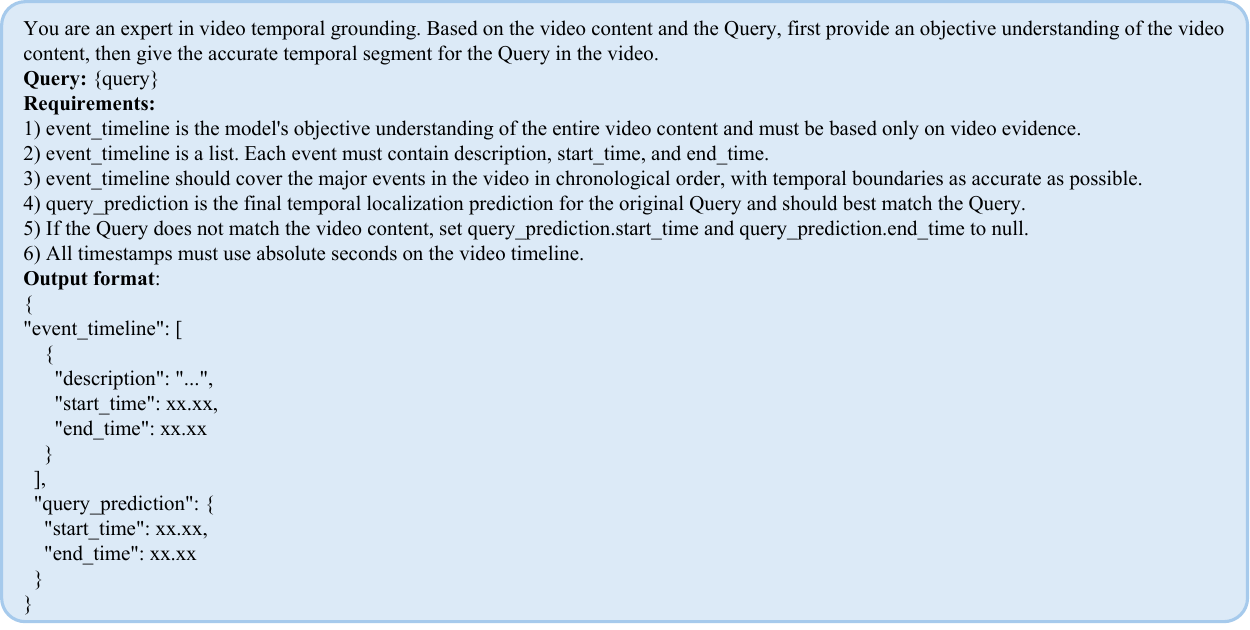}
    \caption{
   The video VLMs take the video and query as input, produce an event-level timeline and a query-specific temporal prediction, and the resulting fields are used for existing-sample filtering and new grounded-sample construction.
    }
    \label{fig:training_data_prompt}
\end{figure*}

\begin{figure*}[!t]
    \centering
    \includegraphics[width=\textwidth]{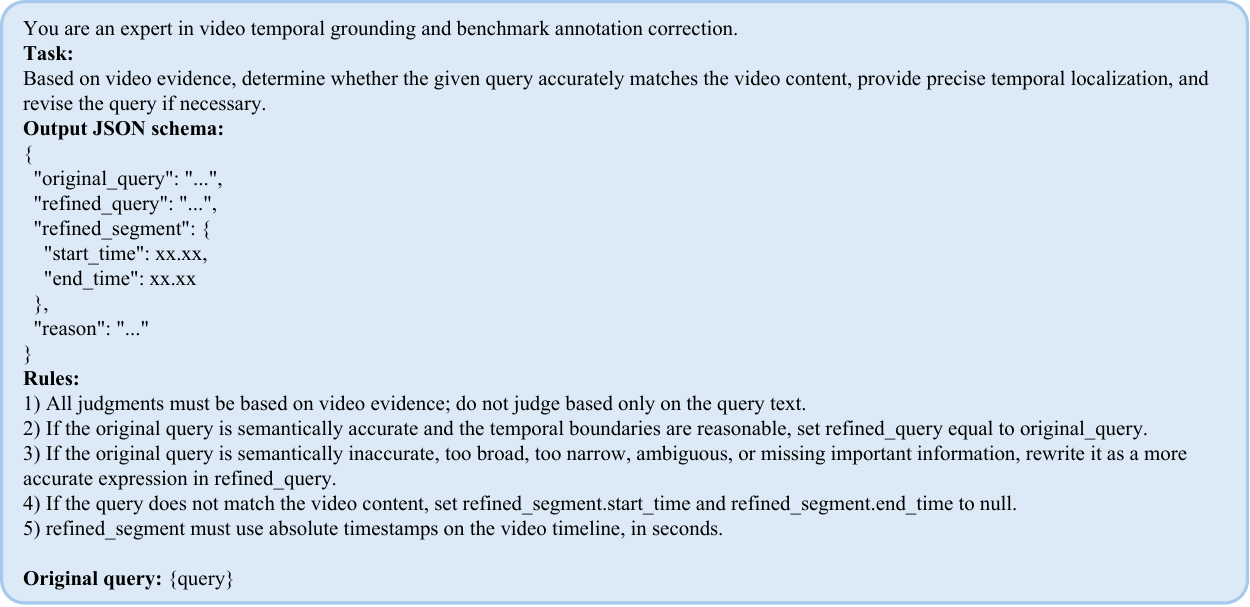}
    \caption{
    Structured prompting protocol for model-assisted benchmark correction. Gemini-3-Pro takes the video, original query, and original temporal annotation as input, and outputs a calibrated query, refined temporal segment, and concise reasoning. These outputs are displayed in the review interface as auxiliary evidence for human correction.
    }
    \label{fig:benchmark_correction_prompt}
\end{figure*}

\begin{figure*}[!t]
    \centering
    \includegraphics[width=\textwidth]{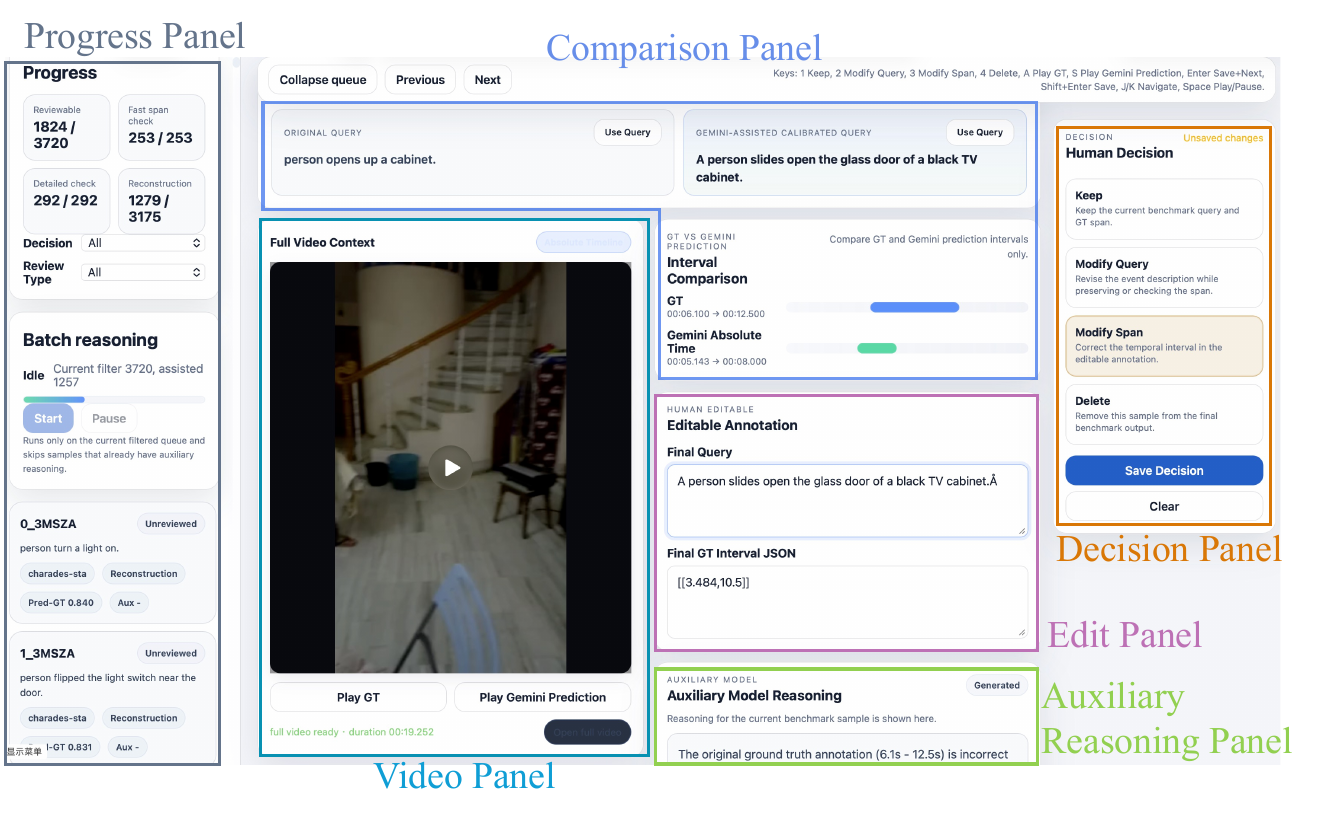}
    \caption{
Web-based interface for benchmark correction. The interface contains a \textit{Progress Panel} for tracking the review queue, a \textit{Comparison Panel} for comparing the original query, Gemini-assisted calibrated query, and temporal intervals, a \textit{Video Panel} for inspecting the full video and localized clips, an \textit{Edit Panel} for modifying the final query and temporal interval, a \textit{Decision Panel} for recording the human decision, and an \textit{Auxiliary Reasoning Panel} for displaying model-generated reasoning.
    }
    \label{fig:benchmark_review_interface}
\end{figure*}

Our data pipeline consists of two complementary branches: training-side data curation and evaluation-side benchmark correction.
The training-side branch uses strong open-source and closed-source video VLMs to filter existing temporal annotations and construct additional grounded samples, which are then converted into the TimePLE supervision format with explicit \texttt{<|TIMESPAN|>} tokens and continuous interval labels.
The evaluation-side branch uses model-assisted human review to correct noisy benchmark annotations, producing a human-verified benchmark for assessing temporally precise grounding.
These two branches serve different purposes: the former improves supervision quality for training, while the latter improves evaluation fidelity.

\subsection{Training Data Curation and Construction}
\label{app:training_data_curation}
We construct the TimePLE training set through two complementary operations:
filtering existing annotated samples and constructing additional grounded samples from model-proposed event candidates.

\paragraph{Data Sources, Video VLMs, and Prompting Protocol}
\label{app:data_sources_prompting}
We build the candidate pool from public video grounding and video-text sources, including InternVid, TimeLens, Ego4D, the Charades-STA training split, and the ActivityNet-Captions training split.
We use Qwen3-VL-30B and Gemini-3-Pro as the open-source and closed-source video VLMs for training data curation.
These models are selected according to our temporal grounding evaluation, where they show the strongest empirical grounding quality among the evaluated candidates.
Their heterogeneous model families also provide complementary candidate intervals for agreement-based filtering and construction.
For training data curation, the video VLMs are prompted to produce structured outputs, including the predicted temporal interval, the corresponding event description, and the query-event correspondence. The complete prompt is illustrated in Fig.~\ref{fig:training_data_prompt}.

\paragraph{Filtering Existing Samples.}
Given an existing annotated sample $(v,q,y)$, where $v$ is the video, $q$ is the query, and $y$ is the original temporal annotation, we ask the video VLMs to predict temporal intervals for the same video-query pair.
A sample is retained when both model predictions reach the filtering IoU threshold with the original annotation.
This step uses video VLM outputs as reliability evidence for existing annotations, rather than as replacements for the original temporal labels.

\paragraph{Constructing New Grounded Samples.}
In addition to filtering existing annotations, we construct new grounded samples from event-level model predictions.
A new sample is accepted when the two video VLMs identify a matched event, where matching requires both sufficient temporal overlap between the predicted intervals and semantic consistency between the event or query descriptions.
This operation expands the training set with additional grounded moments beyond the originally annotated samples.


\paragraph{Conversion to TimePLE Supervision.}
The retained existing samples and newly constructed samples are converted into the same TimePLE supervision format.
Each grounded moment is represented in the textual response by a \texttt{<|TIMESPAN|>} token, and its continuous temporal interval is stored as the corresponding numerical label.
Thus, each \texttt{<|TIMESPAN|>} token is aligned with exactly one temporal interval.


\subsection{Benchmark Correction with Human Review}
\label{app:benchmark_correction}
The evaluation-side branch corrects noisy benchmark annotations through a
model-assisted human review process, where Gemini-3-Pro provides auxiliary
temporal evidence and human annotators determine the final temporal boundaries.

\paragraph{Correction Protocol and Model Assistance.}
For each candidate benchmark sample, the input consists of a video, a natural language query, and the original temporal annotation.
We prompt Gemini-3-Pro to produce auxiliary temporal predictions, including a calibrated query, a refined temporal interval in absolute time, and concise reasoning about the queried event.
Fig.~\ref{fig:benchmark_correction_prompt} illustrates the prompting protocol.
These model outputs are used as auxiliary evidence and are presented to annotators together with the original annotation.


\paragraph{Review Interface and Case Study.}
Fig.~\ref{fig:benchmark_review_interface} shows the web-based review interface and a correction case.
The interface is organized into several functional panels.
The \textit{Progress Panel} tracks the review queue, annotation progress, and the status of the current sample.
The \textit{Comparison Panel} presents the original query, the Gemini-assisted calibrated query, and a visual comparison between the original ground-truth interval and the Gemini-predicted interval.
The \textit{Video Panel} provides the full-video context and localized playback controls for both the original ground-truth span and the Gemini-predicted span.
The \textit{Edit Panel} allows annotators to revise the final query and temporal interval, while the \textit{Decision Panel} records the final human decision as \textit{Keep}, \textit{Modify Query}, \textit{Modify Span}, or \textit{Delete}.
The \textit{Auxiliary Reasoning Panel} displays the model-generated reasoning used as auxiliary evidence during review.
The interface also provides keyboard shortcuts for efficient annotation.
These shortcuts allow annotators to rapidly compare localized clips while still making the final correction through direct video inspection.
In the illustrated case, the original query is \textit{``person opens up a cabinet.''}, while the Gemini-assisted calibrated query specifies the event as \textit{``A person slides open the glass door of a black TV cabinet.''}
The original annotation spans $6.1$--$12.5$ seconds, whereas the Gemini-predicted interval spans $5.143$--$8.0$ seconds.
After comparing the localized clips in the \textit{Video Panel} and inspecting the full-video context, the annotator revises the final interval to $[3.484, 10.5]$.
The final annotation differs from both the original benchmark annotation and the auxiliary Gemini prediction, illustrating that the corrected boundary is determined by human review rather than directly copied from model output.

\begin{figure*}[!t]
    \centering
    \includegraphics[width=\textwidth]{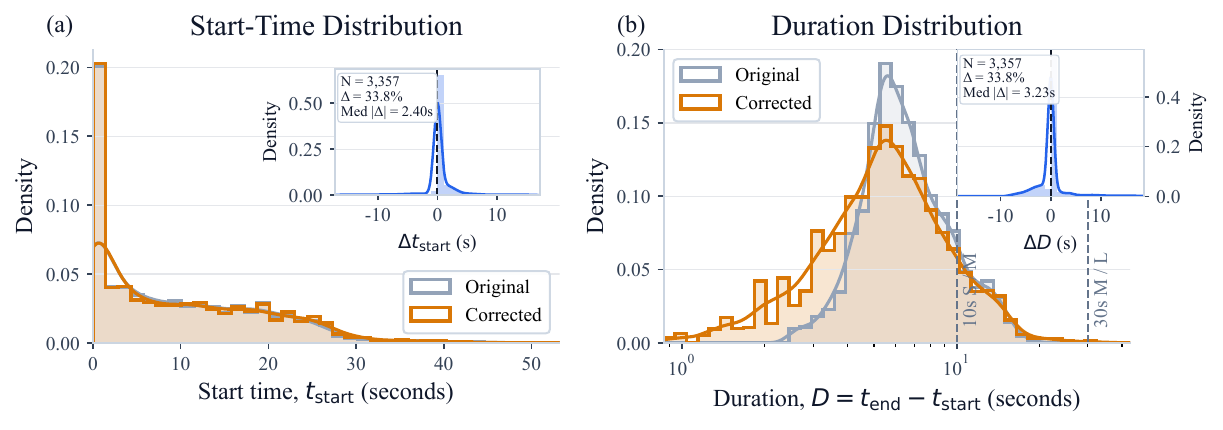}
    \caption{
    Temporal distribution changes after benchmark correction.
    \textbf{(a)} Start-time distributions of the original and corrected annotations, with the inset showing the paired start-time shift.
    \textbf{(b)} Duration distributions of the original and corrected annotations, with the inset showing the paired duration shift.
    Positive shifts indicate later corrected start times or longer corrected durations.
    The dashed vertical lines in the duration plot mark the 10s and 30s thresholds used for duration-stratified evaluation.
    }
    \label{fig:benchmark_correction_distribution}
\end{figure*}

\begin{figure}[!t]
\centering
\begin{minipage}[t]{0.45\textwidth}
    \centering
    \includegraphics[width=\linewidth]{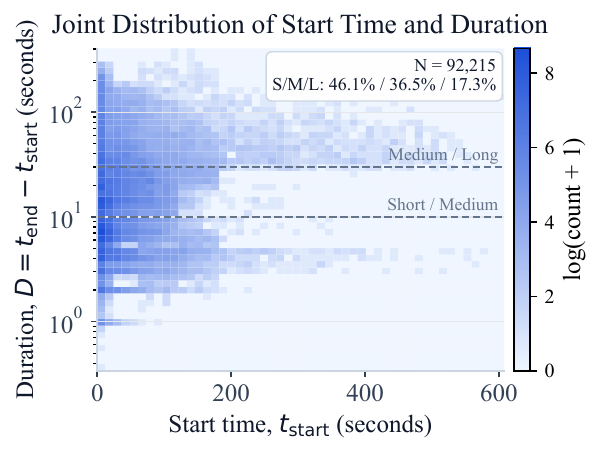}
    \captionof{figure}{
    Temporal distribution of the curated 90K-scale training set.
    The x-axis denotes the ground-truth start time in seconds, and the y-axis denotes the moment duration $D=t_{\mathrm{end}}-t_{\mathrm{start}}$ in seconds.
    The color intensity represents $\log(\mathrm{count}+1)$.
    The duration axis is plotted on a logarithmic scale, and dashed horizontal lines indicate the 10s and 30s thresholds used for duration-stratified evaluation.
    }
    \label{fig:training_data_distribution}
\end{minipage}\hfill
\begin{minipage}[t]{0.5\textwidth}
    \centering
    \includegraphics[width=\linewidth,trim=0 10pt 0 0,clip]{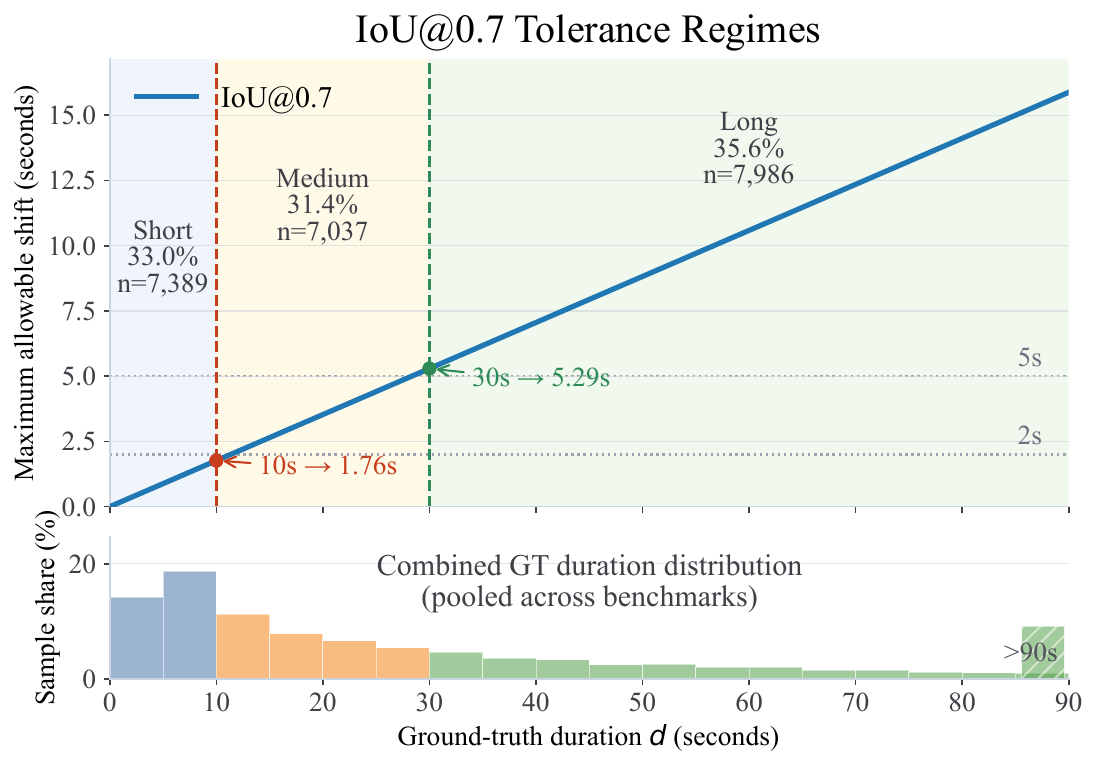}
    \captionof{figure}{Boundary-tolerance regimes and pooled ground-truth duration distribution used for duration-stratified mIoU. The top panel plots the maximum allowable equal-duration temporal shift under IoU@0.7 as a function of ground-truth duration, with vertical dashed lines marking the 10s and 30s thresholds. The bottom panel shows the pooled ground-truth duration histogram across benchmarks using the same Short, Medium, and Long regimes.}
    \label{fig:duration_regime_evidence}
\end{minipage}
\end{figure}

\subsection{Data Statistics}
\label{app:data_statistics}

We further analyze the temporal distributions of the curated training set and
the human-verified benchmark. We focus on two temporal factors: the start time of a grounded moment and its duration.

The former reflects where supervision is distributed along the video timeline, while the latter reflects the diversity of temporal scales covered by the data.
Fig.~\ref{fig:training_data_distribution} shows the joint distribution between ground-truth start time and moment duration in the curated 90K-scale training set.
The heatmap indicates that the training samples cover a broad range oftemporal locations and moment durations, rather than concentrating only on a narrow temporal regime.
The duration axis is shown on a logarithmic scale to make both short and long moments visible, and the dashed horizontal lines mark the 10s and 30s thresholds used for duration-stratified evaluation.

Fig.~\ref{fig:benchmark_correction_distribution} compares the original and
corrected benchmark annotations. The main plots show the global start-time and
duration distributions before and after correction, while the insets show the
paired correction shifts. This visualization captures both the distributional
change induced by benchmark correction and the magnitude of sample-level
boundary adjustments. The corrected annotations preserve the overall temporal
coverage of the benchmark while revising noisy temporal boundaries through
human review.

\section{Evaluation Metric Design}
\label{app:evaluation_metric_design}

\subsection{Standard Temporal Grounding Metrics}
\label{app:standard_temporal_grounding_metrics}
We summarize the standard metrics used for video temporal grounding.
For the $n$-th sample, let the predicted interval be $\hat{I}_n=[\hat{t}^{s}_n,\hat{t}^{e}_n]$ and the ground-truth interval be $I_n=[t^{s}_n,t^{e}_n]$. The temporal intersection and union lengths are
\begin{equation}
\ell_{\cap}(\hat{I}_n,I_n)
=
\max\left(0,\min(\hat{t}^{e}_n,t^{e}_n)-\max(\hat{t}^{s}_n,t^{s}_n)\right),
\end{equation}
\begin{equation}
\ell_{\cup}(\hat{I}_n,I_n)
=
\max(\hat{t}^{e}_n,t^{e}_n)-\min(\hat{t}^{s}_n,t^{s}_n).
\end{equation}
The temporal Intersection-over-Union is
\begin{equation}
\operatorname{IoU}(\hat{I}_n,I_n)
=
\frac{\ell_{\cap}(\hat{I}_n,I_n)}{\ell_{\cup}(\hat{I}_n,I_n)}.
\end{equation}

For a test set with $N$ samples, mean IoU is computed as
\begin{equation}
\operatorname{mIoU}
=
\frac{1}{N}
\sum_{n=1}^{N}
\operatorname{IoU}(\hat{I}_n,I_n).
\end{equation}
mIoU measures the average continuous overlap quality of predicted temporal intervals.

Recall@IoU measures thresholded localization success. Given a threshold $\eta$, it is defined as
\begin{equation}
\operatorname{Recall@}\eta
=
\frac{1}{N}
\sum_{n=1}^{N}
\mathbf{1}
\left[
\operatorname{IoU}(\hat{I}_n,I_n)\geq \eta
\right],
\end{equation}
where common thresholds are $\eta\in\{0.3,0.5,0.7\}$. For methods that output multiple candidate intervals, Recall@K, IoU=$\eta$ is computed as
\begin{equation}
\operatorname{Recall@}K,\operatorname{IoU}=\eta
=
\frac{1}{N}
\sum_{n=1}^{N}
\mathbf{1}
\left[
\max_{1\leq k\leq K}
\operatorname{IoU}(\hat{I}_{n,k},I_n)\geq \eta
\right].
\end{equation}
The single-interval prediction setting corresponds to $K=1$.

\subsection{Duration-Stratified mIoU and Boundary-Tolerance Analysis}
\label{app:duration_stratified_metric}
Aggregate mIoU measures the average overlap quality over all test samples, but it mixes temporal moments with different intrinsic localization difficulty. In particular, short ground-truth intervals are more sensitive to absolute boundary errors: the same temporal shift can severely reduce IoU for a short action while having a much smaller effect on a long event. To make the evaluation more diagnostic, we report duration-stratified mIoU over Short, Medium, and Long ground-truth moments.

The duration buckets are determined by the ground-truth duration rather than the predicted duration, preventing a model from changing its evaluation bucket by altering its prediction length. For the $n$-th sample, let $d_n=t_n^e-t_n^s$ denote the ground-truth duration. We define
\begin{equation}
\begin{aligned}
\mathcal{D}_{S} &= \{\, n : 0 < d_n \leq 10 \,\},\\
\mathcal{D}_{M} &= \{\, n : 10 < d_n \leq 30 \,\},\\
\mathcal{D}_{L} &= \{\, n : d_n > 30 \,\}.
\end{aligned}
\end{equation}
For each bucket $b\in\{S,M,L\}$, the duration-stratified mIoU is
\begin{equation}
\operatorname{mIoU}_{b}
=
\frac{1}{|\mathcal{D}_{b}|}
\sum_{n\in\mathcal{D}_{b}}
\operatorname{IoU}(\hat{I}_n,I_n).
\end{equation}

We use a boundary-tolerance analysis under IoU@0.7 to interpret the 10s and 30s thresholds. Consider a ground-truth interval of duration $d$ and a predicted interval with the same duration but shifted by $\delta$ seconds. Under this equal-duration shift model,
\begin{equation}
\operatorname{IoU}
=
\frac{d-|\delta|}{d+|\delta|}.
\end{equation}
Requiring $\operatorname{IoU}\geq \eta$ gives
\begin{equation}
|\delta|
\leq
d\cdot\frac{1-\eta}{1+\eta}.
\end{equation}
For $\eta=0.7$, this becomes $|\delta|\leq 0.1765d$. Therefore, a 10-second moment allows only approximately 1.76s shift error, while a 30-second moment allows approximately 5.29s.

Figure~\ref{fig:duration_regime_evidence} combines this boundary-tolerance curve with the pooled ground-truth duration distribution across benchmarks. The three regimes contain comparable portions of the evaluation samples, with 33.0\% Short, 31.4\% Medium, and 35.6\% Long moments. Thus, the proposed duration split is both interpretable under IoU@0.7 and empirically meaningful for evaluating temporal grounding models across samples with different precision requirements.








\section{Additional Exploration of Progressive Interval Optimization}
\label{app:progressive_interval_analysis}
\subsection{Motivation and Setup}
\label{app:post_sft_motivation}
Post-SFT optimization has become a common strategy for improving LLMs and VLMs beyond supervised fine-tuning, typically by using reward-guided policy updates, on-policy distillation, or response-level preference signals.
This paradigm is also attractive for video temporal grounding, where the final evaluation depends on the quality of the interval decoded from generated responses.
We therefore explore whether TimePLE can benefit from additional post-SFT optimization after the interval-supervised SFT stage.

However, post-SFT optimization for TimePLE is different from ordinary response-level optimization. TimePLE (SFT) already provides direct supervision to the generated \texttt{<|TIMESPAN|>} token through interval distributional loss, overlap loss, and boundary loss.
As shown in the main ablation, applying a standard GRPO-style objective with commonly used VTG rewards does not further
improve over TimePLE (SFT).
This suggests that outcome-level rewards computed from the final decoded interval may provide limited additional guidance once the latent interval interface has already been aligned by span-level supervision.

Motivated by this observation, we explore two interval-aware post-SFT strategies:
\textit{Counterfactual Span Distribution Optimization} (CSDO) and \textit{Trust-Region Span Posterior Distillation} (TR-SPD).
CSDO constructs a reward-improved target distribution over the canonical span grid by evaluating counterfactual movements around the old-policy prediction,
while TR-SPD constructs posterior span targets under a trust-region constraint relative to the self-prior and distills the current span distribution toward accepted targets.
All variants are initialized from the same TimePLE (SFT) checkpoint and trained with the same data, video processing settings, and evaluation protocol.
The interval decoder is frozen during post-SFT optimization, so the additional objectives mainly affect the VLM hidden representation and span interface.

\subsection{Explored Post-SFT Training Algorithms}
\label{app:post_sft_algorithms}

We explore two post-SFT training algorithms that operate directly on the canonical span distribution associated with the generated \texttt{<|TIMESPAN|>} token.
Both methods keep the interval decoder frozen and optimize the actor through the span distribution predicted from the current VLM hidden state.
They differ in how the post-SFT target distribution is constructed.

\paragraph{Counterfactual Span Distribution Optimization.}
CSDO augments the response-level policy update with an auxiliary
span-distribution objective.
For each rollout sample, we first obtain a detached old-policy span distribution $p_{\mathrm{old}}$ and its corresponding span
feature at the \texttt{<|TIMESPAN|>} position.
The old-policy distribution is used to compute an expected canonical coordinate $\mu_{\mathrm{old}}$.
Then, for each canonical grid cell $c$ with coordinate $x_c$, we construct a local counterfactual coordinate by moving a small step from the old expectation toward that cell:
\begin{equation}
\mu_{+}(c) = (1-\eta)\mu_{\mathrm{old}} + \eta x_c .
\end{equation}
Using the detached old-policy span feature and the frozen duration-adaptive decoder, each counterfactual coordinate is decoded into a temporal interval and scored against the ground-truth span.
The resulting reward improvement
\begin{equation}
A(c) =
R(\mathrm{decode}(\mu_{+}(c))) -
R(\mathrm{decode}(\mu_{\mathrm{old}}))
\end{equation}
measures whether moving probability mass toward cell $c$ would improve the decoded interval.
We then form a reward-improved target distribution over the canonical span grid:
\begin{equation}
q(c) =
\mathrm{softmax}\left(
\log p_{\mathrm{old}}(c) + \frac{\widetilde{A}(c)}{\tau}
\right),
\end{equation}
where $\widetilde{A}(c)$ denotes the normalized counterfactual advantage.
The current actor distribution $p_{\mathrm{cur}}$ is trained to match this detached target through a cross-entropy loss, optionally regularized by a span-level KL term to the reference distribution.

This design aims to provide a more geometry-aware signal than scalar response-level rewards.
Instead of assigning a single reward to the final decoded interval, CSDO estimates which regions of the canonical span grid would locally improve the decoded interval and uses this information to reshape the current span distribution.
However, the supervision is still derived from outcome-level interval rewards and therefore provides only indirect credit assignment to the hidden-state geometry.

\paragraph{Trust-Region Span Posterior Distillation.}
TR-SPD follows a different post-SFT strategy.
Rather than constructing a local counterfactual target during each actor update, it builds posterior span targets from the old-policy span distribution under a self-prior trust-region constraint.
Let $p_b$ denote the detached old-policy span distribution used as the self-prior.
For a set of temperature candidates, TR-SPD constructs posterior candidates $q_{\tau}$ and accepts a candidate only when its deviation from the self-prior satisfies
\begin{equation}
D_{\mathrm{KL}}(q_{\tau}\,\|\,p_b) \leq \rho .
\end{equation}
If no posterior candidate satisfies the trust-region budget, the sample falls back to weak self-prior retention.
The accepted posterior target is then used as a detached teacher distribution for span-level distillation.

In the online training stage, TR-SPD optimizes the current span distribution toward the accepted posterior target while using text-level KL regularization to preserve the response behavior.
Unlike CSDO, this algorithm does not rely on the standard response-level policy-gradient loss as the main optimization signal.
Instead, it treats post-SFT training primarily as a posterior distillation problem over the canonical interval grid.
The trust-region constraint is introduced to avoid moving the teacher distribution too far from the already learned TimePLE (SFT) span prior, while still allowing reward-improving posterior targets to guide the actor.

The two algorithms therefore represent complementary attempts to improve TimePLE after interval-supervised SFT.
CSDO performs online counterfactual reward shaping over the span grid, whereas TR-SPD performs trust-region posterior distillation from self-prior span distributions.
Both are specifically designed for the interval-native prediction space, but neither assumes that ordinary response-level rewards alone are sufficient to refine the latent
\texttt{<|TIMESPAN|>} representation.

\subsection{Experimental Results}
\label{app:post_sft_results}
\begin{table}[t]
\centering
\caption{
Post-SFT optimization results initialized from the same TimePLE (SFT) checkpoint.
C-STA, A-Net, QVH, and C-TPLE denote Charades-STA, ActivityNet-Captions,
QVHighlights, and Charades-TimePLE, respectively.
Numbers in parentheses report mIoU changes relative to TimePLE (SFT).
}
\label{tab:post_sft_optimization}
\fontsize{8}{9.6}\selectfont
\setlength{\tabcolsep}{3.2pt}
\renewcommand{\arraystretch}{1.05}
\begin{tabular}{@{}l|cccc|c@{}}
\toprule
Training setting
& C-STA & A-Net & QVH & C-TPLE & Avg. \\
\midrule
\rowcolor{cyan!8}
TimePLE (SFT)
& \textbf{57.2} & \textbf{49.6} & \textbf{65.3} & \textbf{63.4} & \textbf{58.9} \\

\qquad + GRPO
& 57.2 \same{0.0}
& 49.6 \same{0.0}
& 65.1 \dropv{0.2}
& 63.3 \dropv{0.1}
& 58.8 \dropv{0.1} \\

\qquad + CSDO
& 56.9 \dropv{0.3}
& 48.9 \dropv{0.7}
& 63.4 \dropv{1.9}
& 62.5 \dropv{0.9}
& 57.9 \dropv{1.0} \\

\qquad + TR-SPD
& 57.2 \same{0.0}
& 48.4 \dropv{1.2}
& 63.4 \dropv{1.9}
& 62.9 \dropv{0.5}
& 58.0 \dropv{0.9} \\

\bottomrule
\end{tabular}
\end{table}

Table~\ref{tab:post_sft_optimization} reports the results of the explored post-SFT optimization strategies.
The results show that none of the post-SFT methods provides a stable improvement over the interval-supervised SFT checkpoint.
Standard GRPO remains nearly unchanged, while CSDO and TR-SPD lead to small performance drops on average despite their interval-aware target construction.

Overall, the post-SFT results support the observation from the main ablation: after interval-supervised SFT, additional reward-style or posterior-distillation training does not reliably improve TimePLE.
Although CSDO and TR-SPD introduce span-distribution targets that are more aligned with the canonical interval
space than ordinary response-level rewards, their improvements do not transfer to consistent benchmark-level gains.

\subsection{Mechanistic Analysis of Post-SFT Optimization}
\label{app:post_sft_discussion}

The post-SFT results suggest that the main optimization challenge in TimePLE is not response selection, but latent interval geometry.
In conventional instruction tuning or reward-guided VLM training, the optimized object is usually close to the model's native output space: token probabilities, response preferences, or scalar rewards assigned to complete generations.
In contrast, TimePLE localizes a moment through the hidden state of a generated \texttt{<|TIMESPAN|>} token, which is decoded into a distribution over the canonical interval space.
Therefore, effective optimization must shape not only the generated response format, but also the geometry of the span-token representation before decoding.

This explains why interval-supervised SFT remains the strongest training signal in our pipeline.
During TimePLE (SFT), the \texttt{<|TIMESPAN|>} hidden state is directly aligned with ground-truth interval distributions through distributional, overlap, and boundary losses. These losses provide dense supervision in the same interval space used by the decoder.
By contrast, CSDO and TR-SPD construct post-SFT targets from the model's own old-policy span distribution.
CSDO moves probability mass according to counterfactual decoded reward improvements, while TR-SPD distills posterior targets constrained by a self-prior trust region.
Although both targets are more interval-aware than ordinary response-level rewards, they are still bootstrapped from the current model's span prior and decoded outcomes rather than from new external interval evidence.

The limited gains therefore indicate a structural limitation of the explored post-SFT objectives.
If the target remains too close to the self-prior, it adds little information beyond the already supervised SFT solution.
If the target moves too far from the self-prior, it can disturb the learned interface between language generation and interval decoding.
This tension is especially important for TimePLE because the interval decoder is frozen during post-SFT training, so all additional optimization must be absorbed by the VLM hidden representation and span interface.
The results suggest that further improvement may require post-SFT signals that are denser and less self-referential, such as correction-aware interval distillation, external teacher distributions, or geometry-consistent supervision that directly constrains the canonical interval distribution.

\begin{findingbox}
Post-SFT optimization for TimePLE is fundamentally different from ordinary
reward-guided response optimization. Since the prediction is mediated by the
latent \texttt{<|TIMESPAN|>} representation, the most effective signal is one
that directly supervises the canonical interval distribution. CSDO and TR-SPD
make the post-SFT objective more interval-aware, but their targets are still
derived from the model span prior and decoded outcomes. This explains why they
do not consistently improve over interval-supervised SFT and suggests that
future post-SFT training should rely on denser, externally grounded, and
geometry-consistent interval targets.
\end{findingbox}


\end{document}